\documentclass{article}

\PassOptionsToPackage{numbers, compress}{natbib}

\usepackage[preprint]{neurips_2026}

\usepackage[utf8]{inputenc} 
\usepackage[T1]{fontenc}    
\usepackage[hidelinks]{hyperref} 
\usepackage{url}            
\usepackage{booktabs}       
\usepackage{amsmath}        
\usepackage{amsfonts}       
\usepackage{capt-of}        
\usepackage{enumitem}       
\usepackage{listings}       
\usepackage{nicefrac}       
\usepackage{microtype}      
\usepackage{xcolor}         
\usepackage{graphicx}
\usepackage{multirow}
\usepackage{arydshln}

\definecolor{customgreen}{RGB}{0,128,0}
\definecolor{lightblue}{RGB}{70,130,180}

\newcommand{\best}[1]{\textbf{\textcolor{customgreen}{#1}}}
\newcommand{\second}[1]{\textbf{\textcolor{lightblue}{#1}}}
\newcommand{\NA}{--}
\newcommand{\fr}[1]{\,#1}

\title{3D-CoS: A New 3D Reconstruction Paradigm \\Based on VLM Code Synthesis}

\author{%
  \normalfont
  \parbox{0.94\textwidth}{\centering
    \textbf{Yuhao Wang\textsuperscript{*1},
    Puyi Wang\textsuperscript{*2},
    Linjie Li\textsuperscript{3},
    Zhengyuan Yang\textsuperscript{3}}\\[0.25em]
    \textbf{Kevin Qinghong Lin\textsuperscript{4},
    Yu Cheng\textsuperscript{2}}\\[0.85em]
    \begin{tabular}{c}
      \textsuperscript{1}Shanghai Jiao Tong University\\
      \textsuperscript{2}The Chinese University of Hong Kong\\
      \textsuperscript{3}Microsoft \quad
      \textsuperscript{4}University of Oxford\\
      \textsuperscript{*}Equal contribution.
    \end{tabular}
  }
}

\begin{document}

\maketitle

\begin{abstract}
Most recent 3D reconstruction and editing systems operate on implicit and explicit representations such as NeRF, point clouds, or meshes. While these representations enable high-fidelity rendering, they are fundamentally low-level and hard to control programmatically. In contrast, we propose and systematically evaluate a new 3D reconstruction paradigm, 3D Code Synthesis (3D-CoS), where 3D assets are constructed as executable Blender code, a programmatic and interpretable medium. To assess how well current VLMs can use code to represent 3D objects, we evaluate representative open-source and closed-source VLMs in code-based reconstruction under a unified protocol. We further introduce a suite of structured code-synthesis workflows, including blueprint-based planning, Retrieval-Augmented Generation (RAG) over Blender API documentation, few-shot geometric demonstrations, and a component-level Agent workflow for part-wise code generation. To demonstrate the unique advantages of this representation, we further evaluate localized text-driven modifications and compare our code-based edits with a point-cloud-based 3D editing baseline. Our study shows that code as a 3D representation offers strong controllability and locality, yielding stronger edit fidelity and better preservation of unedited regions in our targeted editing evaluation. Our work also analyzes the potential of this paradigm, delineates the current capability frontier of VLMs for programmatic 3D modeling, and highlights code synthesis as a promising direction for editable 3D reconstruction. 
\end{abstract}

\section{Introduction}
Recent breakthroughs in large vision-language models (VLMs) have led to remarkable progress in multimodal understanding, logical reasoning, and tool usage. These models have shown the ability to operate within a ``perception, reasoning, planning, and execution" loop, and automatically generate executable code to accomplish complex tasks~\citep{gao2023palprogramaidedlanguagemodels, li2024chaincodereasoninglanguage, liang2023codepolicieslanguagemodel}. This capability suggests a new path for 3D: instead of recovering geometry purely as meshes, point clouds, or implicit fields, we can generate executable programs that reconstruct 3D assets inside 3D engines (e.g., Blender~\citep{blender}, Unity~\citep{unity_engine}, Unreal~\citep{unreal_engine}). Code as a 3D representation brings interpretability, editability, and compositional control. 3D components are constructed in an explicit and parameterized manner, and are verifiable by execution. Moreover, the mature APIs of 3D engines \citep{blender_api_doc, unreal_cpp_api, unity_scripting_api} enable full programmatic control over creation and rendering, providing a robust substrate for automation~\citep{blendermcp}.

\begin{figure}
    \centering
    \includegraphics[width=\textwidth]{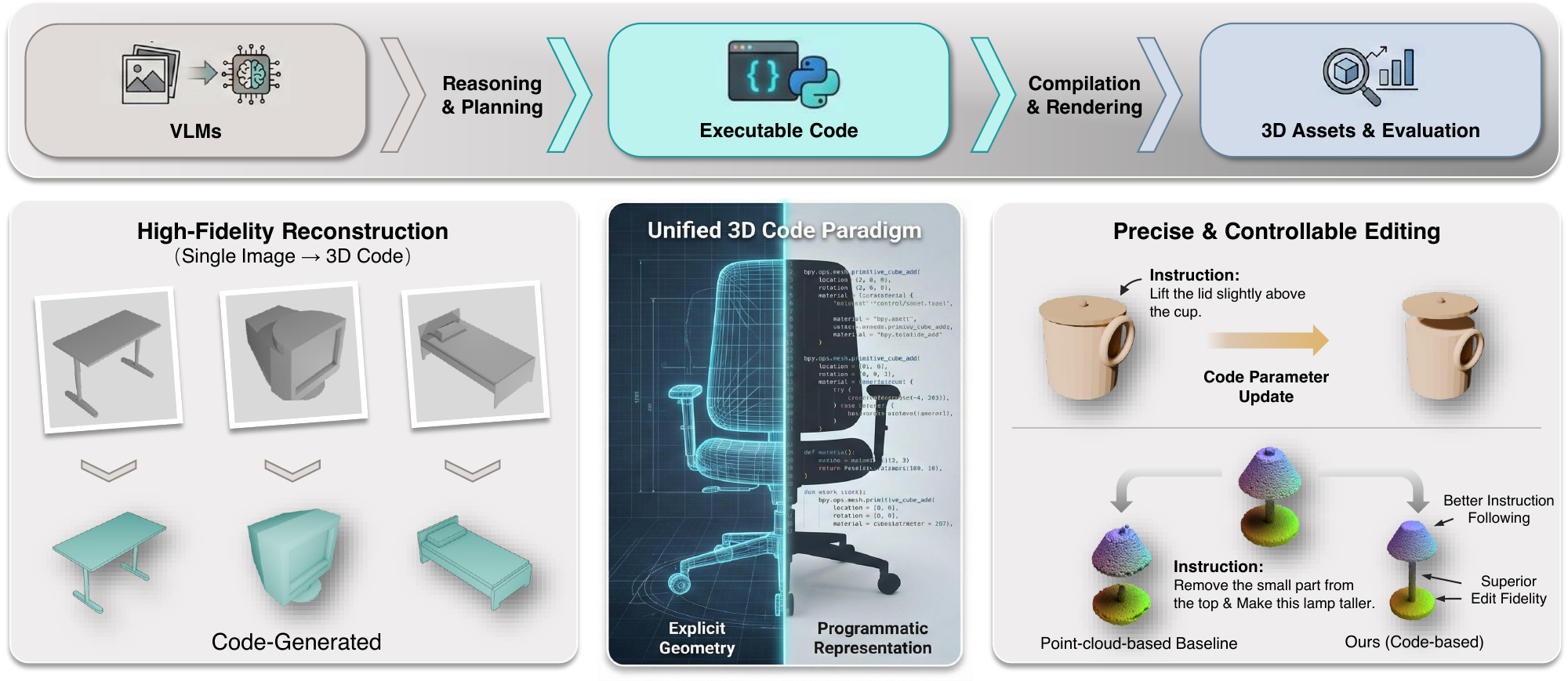}
    \caption{\textbf{An overview of our 3D code modeling paradigm.} The top workflow summarizes our core process: code synthesis via VLMs, and its subsequent evaluation. Our work treats code as a unified representation for 3D assets. (Left) We demonstrate its capability in \textbf{reconstruction}, generating high-fidelity objects from a single \textbf{image}. (Right) We highlight its advantages in \textbf{editing}, where code-driven edits achieve superior fidelity compared to a point-cloud-based 3D editing baseline.}
    \label{fig:teaser}
    \vspace{-16pt}
\end{figure}

Several recent works have explored the feasibility of using code to generate and edit 3D assets, and two recent lines of work motivate our study. LL3M~\citep{lu2025ll3m} demonstrates text-driven 3D asset creation by coordinating agents that write Blender scripts, evidencing that code can serve as a powerful representation for modeling geometry, layout, and appearance. BlenderGym~\citep{gu2025blendergym} introduces a benchmark that tasks VLM systems with code-based 3D scene editing and shows that state-of-the-art models can comprehend programmatic code and further make targeted code-level modifications. Together, these works validate the feasibility of ``code for 3D" while revealing a gap in image-conditioned reconstruction and in systematic evaluation specific to 3D reconstruction. In reconstruction, the image input is essential: it supplies silhouette constraints, object pose, and disambiguates topology and fine details that text alone cannot specify. Equally important is a standardized pipeline to evaluate code-as-representation under image conditioning. We fill this gap with a systematic study of image-conditioned, code-based 3D reconstruction, accompanied by a unified evaluation protocol across multiple VLM families.

We focus on the image $\rightarrow$ code $\rightarrow$ 3D setting and address two fundamental questions: Why use code as the representation? and What is the reconstruction ceiling? First, unlike purely implicit (e.g., NeRF, Gaussian Splatting) or low-level explicit representations, code enables fine-grained control and reliable iteration through high-level logic. Second, regarding the ceiling, we observe that the vast majority of human-made 3D assets originate from programmatic workflows. Datasets like Fusion 360~\citep{willis2020fusion} capture these parametric design histories, and even large-scale mesh collections like ModelNet~\citep{modelnet_wu20153dshapenetsdeeprepresentation} and ShapeNet~\citep{chang2015shapenetinformationrich3dmodel} are derived from CAD models. Therefore, the theoretical ceiling of this paradigm is the recovery of the original executable design program used by human modelers.

In this work, we propose a novel paradigm for 3D reconstruction using programmatic code on the Blender platform~(Figure~\ref{fig:teaser}). This code-based representation makes results easier to edit, offers finer-grained control, and is more interpretable than standard mesh formats. To systematically evaluate the capabilities of modern VLMs in this setting, we introduce a code-based reconstruction benchmark that evaluates representative open- and closed-source VLMs on single-image reconstruction under a unified evaluation protocol, analyzes four direct prompting paradigms together with an Agent workflow evaluated on closed-source models, and compares their results to representative external 3D reconstruction baselines. We report standard 3D geometric metrics and complementary 2D metrics computed from the same camera view used to generate the input condition image. These 2D metrics evaluate how well a model reconstructs the observed view when it receives only a single-view input.
Beyond reconstruction, we include a code-based editing protocol to expose the unique strengths of programmatic control (e.g., targeted parameter changes, retention of unedited areas) relative to point-cloud-based 3D editing pipelines, further highlighting the potential of code as an editable 3D representation. 

Our main contributions are threefold:
\begin{itemize}
\item We introduce a benchmark protocol and metrics suite for Blender-code reconstruction, and use it to evaluate representative open- and closed-source VLMs in terms of their code-based 3D reconstruction capabilities.
\item We evaluate four direct prompting paradigms and a component-level Agent workflow for VLM-based code synthesis, showing that structured workflows achieve promising reconstruction quality and that the best-performing Gemini configurations approach InstantMesh on specific metrics.
\item We demonstrate that the proposed code-based representation offers substantial advantages for 3D editing, and empirically validate its benefits against a point-cloud-based 3D editing baseline through targeted editing evaluations.
\end{itemize}

\section{Related Work}

\noindent \textbf{Traditional 3D Reconstruction Representations.}
Existing methods mainly develop along two lines: (i) \emph{Neural or continuous rendering representations} such as neural-radiance-field-based methods~\citep{mildenhall2021nerf, poole2022dreamfusion,kosiorek2021nerf,wang2023rodin,wang2022clip}, point-based 3D Gaussian Splatting methods~\citep{kerbl2023_3dGaussian,chen2024text2gaussian,yi2024gaussiandreamer,wu2025textsplat}, and other approaches that learn a latent space and decode it into neural shape representations~\citep{zhang20233dshape2vecset,jun2023shap,lan2024ln3diff}. This family excels in visual fidelity, but typically offers limited precise control, lacks interoperability with standard graphics pipelines, and often relies on heavy optimization or bespoke training.
(ii) \emph{Explicit shape representations} (point clouds, voxels, meshes) are more amenable to geometric measurement and integration with existing engines, and have been extensively studied~\citep{chen2021decor,li2021sp,ibing20213d,vahdat2022lion}. However, they lack shared high-level primitives and constraints, making automated control and cross-category, generalizable editing challenging.

\noindent \textbf{Code Based 3D Representations.}
Regarding Domain-Specific Language (DSL) approaches, 3D Shape Programs~\citep{tian2018learning} encode repeated and symmetric structures as programs. ShapeAssembly~\citep{jones2020shapeAssembly} designs DSLs specifically for 3D shape structures, constructing hierarchical and reconfigurable shape programs. 
These approaches often rely on custom DSLs whose primitive vocabularies and domain-specific grammars constrain free-form geometry. 
In the CAD-oriented line of code-based representations, DeepCAD~\citep{Wu_2021_ICCV} treats CAD operation sequences (sketch, extrude, etc.) as program sequences. Later text- and point-to-CAD methods~\citep{xu2024CADMLLM} similarly exploit parametric operation sequences to obtain editable designs.
These methods typically rely on custom DSLs or commercial CAD environments, which are best suited to constructing regular geometries for downstream manufacturing rather than the free-form, animation-oriented content creation enabled by tools like Blender.

\noindent \textbf{Large Models for 3D Generation and Editing via Code.}
The success of Large Models (LMs) in leveraging code to solve problems~\citep{gao2023palprogramaidedlanguagemodels} has inspired exploration into using LMs to generate code for manipulating 3D objects.
BlenderAlchemy~\citep{huang2024blenderalchemy} generates materials in Blender for existing geometry. SceneCraft~\citep{hu2024scenecraft} retrieves 3D assets and employs an LLM to organize them into a coherent spatial layout. 
LL3M~\citep{lu2025ll3m} generates 3D assets from text guidance, incorporating geometry and appearance attributes and BlenderMCP~\citep{blendermcp} uses a single LLM calling Blender functions via the Model Context Protocol~\citep{mcp2024}. BlenderGym~\citep{gu2025blendergym} utilizes VLMs for 3D scene reconstruction through code editing. MeshCoder~\citep{dai2025meshcoder} fine-tunes an LLM to translate 3D point clouds into editable Blender scripts. However, these methods either do not directly compare code with traditional 3D representations in terms of reconstruction and editing behavior, or do not systematically evaluate image-conditioned 3D reconstruction through executable code.

In contrast, we demonstrate the benefits of code compared to traditional representations and provide a comprehensive evaluation of the capability of current VLMs to reconstruct and edit 3D objects directly from image inputs.

\section{Method}
\subsection{3D Reconstruction Paradigms for Code Synthesis}
\label{sec:reconstruction}

\textbf{Problem setup.} Given a single image $\mathcal{I}$, we generate an executable Blender Python script whose rendered 3D object matches the target object as closely as possible. To study how different code synthesis paradigms affect reconstruction, we compare four direct prompting paradigms, \emph{Single-call}, \emph{Planning}, \emph{RAG}, and \emph{Few-shot}, together with an \emph{Agent Paradigm} for component-level generation. We also design a text-conditional reconstruction variant described below.

\begin{figure*}[t]
    \centering
    \includegraphics[width=\linewidth]{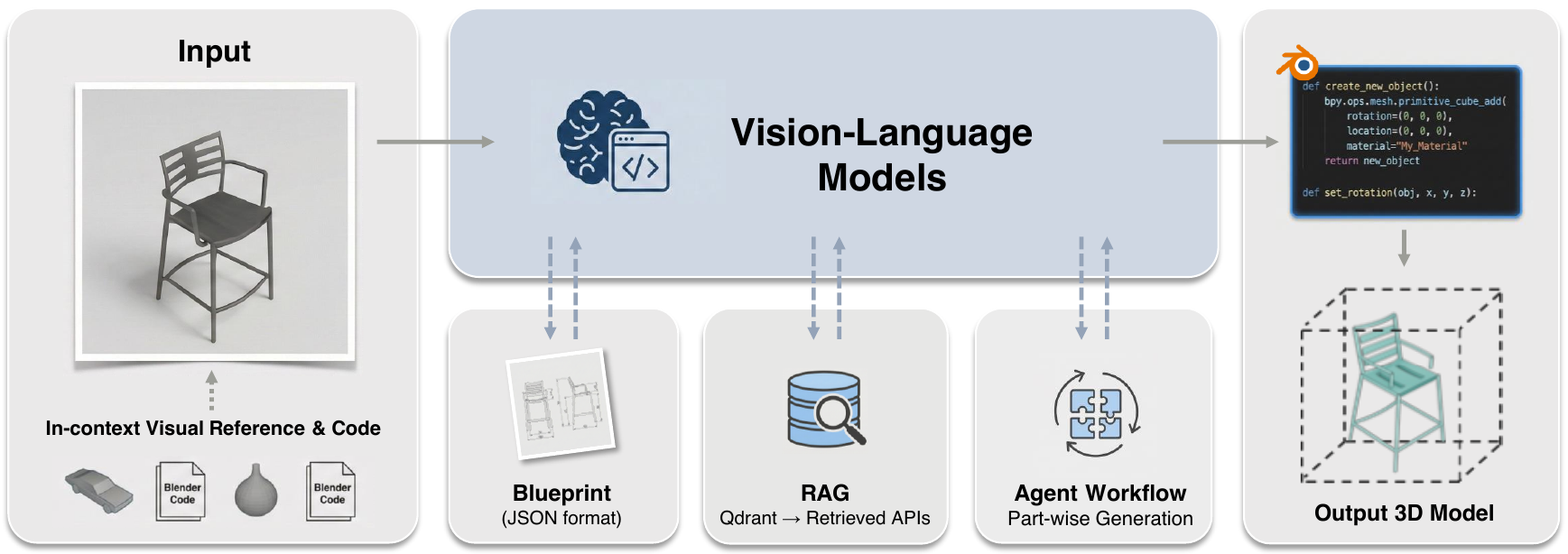}
    \caption{\textbf{Overview of our VLM-based 3D reconstruction pipeline.} Solid arrows show the image-to-code-to-3D workflow; dashed arrows show optional context from visual-code examples, blueprints, RAG-based API retrieval, and part-wise Agent generation. Appendix~\ref{sec:detail_artifacts} provides example artifacts.}
    \label{fig:reconstruction_pipeline}
    \vspace{-10pt}
\end{figure*}

\textbf{Single-call Paradigm.}
We prompt the VLM once to reconstruct the object and output a complete script, requiring the script to encapsulate all reconstruction logic in one callable function with explicit parameters as input. This is the fastest setting, but it often misses fine details or mis-specifies parameters for complex shapes.

\textbf{Planning Paradigm.}
We decouple \emph{what to build} from \emph{how to implement it}: Stage~1 infers a quantitative blueprint from $\mathcal{I}$, and Stage~2 generates code conditioned on that blueprint.

\textit{Stage 1: quantitative blueprint.} The VLM defines an object-level base dimension (e.g., overall width or height) and sets it to $1.0$, then specifies the dimensions of all parts in $\mathcal{B}$ as ratios to this base. We also require a feasibility check: under occlusion or perspective ambiguity, the model may propose minimal adjustments to ensure structural stability. See Appendix~\ref{sec:detail_artifacts} for a blueprint example.

\textit{Stage 2: code generation.} Conditioned on $(\mathcal{I}, \mathcal{B})$, the VLM generates a script that preserves the numeric structure in $\mathcal{B}$. The image is used mainly for non-parametric details (e.g., complex curves), with only minimal edits allowed to resolve infeasible geometry. Compared to Single-call, Planning improves interpretability via an explicit blueprint and a clean separation between geometry and API usage.

\textbf{RAG Paradigm.}
Relying solely on internal knowledge can lead to hallucinated API calls and make the model overlook appropriate but long-tail functions, so we retrieve Blender~4.4 documentation as external context during code synthesis. We crawl the Blender~4.4 docs using the Sphinx inventory~\citep{sphinx_doc} and sphobjinv~\citep{skinn_sphobjinv_2024}, extract functions, signatures, and parameter semantics from \texttt{bpy}, \texttt{bmesh}, and \texttt{mathutils}, and store each entry as a standardized JSON item $\mathcal{D}$ (Listing~\ref{supple:RAG database entry}). The resulting database covers 1,683 pages and 21,102 functions, which we embed and index in Qdrant~\citep{qdrant2025github} for hybrid retrieval.

We first infer the same quantitative blueprint $\mathcal{B}$ as in the Planning paradigm. For each component in $\mathcal{B}$, the VLM generates API-oriented queries with module preferences and keyword constraints, mapping semantic parts to primitives and operations such as bevel or solidify (Listing~\ref{supple:vlm_query}). We retrieve Top-$k$ ($k=8$) documentation chunks, consolidate them into background knowledge $\mathcal{K}$ in a \emph{component $\rightarrow$ candidate-API list} schema (Listing~\ref{supple:rag_example}), and generate the final script conditioned on $(\mathcal{I}, \mathcal{B}, \mathcal{K})$.

\textbf{Few-shot Paradigm.}
To probe in-context learning capabilities of VLMs, we prepend a small set of exemplary pairs $(\mathcal{I}_{ex}, \mathcal{S}_{ex})$ to the task instruction. We generate the exemplar scripts with Gemini 3 Pro on samples disjoint from ModelNet10 and manually curate them to ensure consistent blueprint-style reasoning and correct API usage. By observing exemplary pairs, the model is guided to implicitly learn the underlying geometric reasoning patterns and syntactic standards, applying them to the target reconstruction task.

\textbf{Agent Paradigm.}
Besides the four direct prompting paradigms above, we study an agent paradigm that introduces a component-level reconstruction workflow. Given an input image $\mathcal{I}$, the agent first constructs a structured object plan $\mathcal{P}=\{p_i\}_{i=1}^{N}$, where each part $p_i$ describes a major component together with its approximate geometry and spatial placement. Conditioned on $\mathcal{P}$, the model synthesizes \texttt{bpy} code for each part separately, so local structure is specified explicitly rather than being produced in a single global pass. The generated code is executed during construction, and when a part program fails, the agent revises that part and executes the revised program before continuing. After all parts are completed, the part-level programs are assembled into a full reconstruction, yielding the final 3D model. This design is particularly suited to improving local geometric fidelity and more balanced recovery of fine object components.

\textbf{Variant: From Reconstruction to Editing.}
This variant extends reconstruction to incorporate an edit intent $\mathcal{T}_{edit}$. Reusing the Planning workflow, Stage~1 predicts an edited blueprint $\mathcal{B}_{edit}$ from $(\mathcal{I}, \mathcal{T}_{edit})$, and Stage~2 generates the final script conditioned on $(\mathcal{I}, \mathcal{B}_{edit})$. This variant demonstrates the flexibility of our paradigm by unifying reconstruction and editing into a single generation process.

\subsection{Reconstruction Benchmark}
\label{sec:recon_benchmark}
We introduce a unified benchmark for code-based 3D reconstruction that combines explicit geometric registration with complementary 3D and 2D evaluation. The protocol is motivated by a characteristic ambiguity of executable 3D code: a generated asset may preserve the target structure while differing in global scale or pose convention. The benchmark therefore factors out such nuisance transformations while retaining sensitivity to geometric fidelity and view-consistent reconstruction quality.

\textbf{Registration protocol.}
\label{sec:registration}

Before metric computation, we apply an upright-preserving similarity alignment. Global scale, translation, and yaw are optimized, whereas pitch and roll are fixed. We first enumerate candidate yaw angles and score them with a robust symmetric surface-distance objective, then optimize the best candidates with iterative closest point updates under the same yaw-only constraint. The selected transform is applied to the generated mesh for all quantitative evaluation.

\textbf{Benchmark scope.}
The benchmark is instantiated on a controlled single-view reconstruction setting derived from ModelNet10 and includes an easy/hard split for analyzing sensitivity to structural complexity. We further extend this setting with a text-conditional reconstruction variant, denoted ModelNet10-V, to qualitatively assess generation conditioned jointly on an input image and an edit instruction. For ModelNet10 reconstruction, quality is assessed through complementary 3D metrics of geometric accuracy and structural coverage together with 2D metrics computed from the input-condition camera view. Detailed dataset construction and metric definitions are provided in the experimental setup.

\subsection{3D Editing Paradigm with Code Modification}
\label{sec:editing}
A key reason we choose code as the representation for 3D shapes is the programmatic controllability it provides for subsequent editing operations. When a 3D object generated by code needs to be modified, we can make adjustments directly at the code level, leveraging the VLM's powerful comprehension and reasoning capabilities.

\begin{figure*}[t!]
    \centering
    \includegraphics[width=\linewidth]{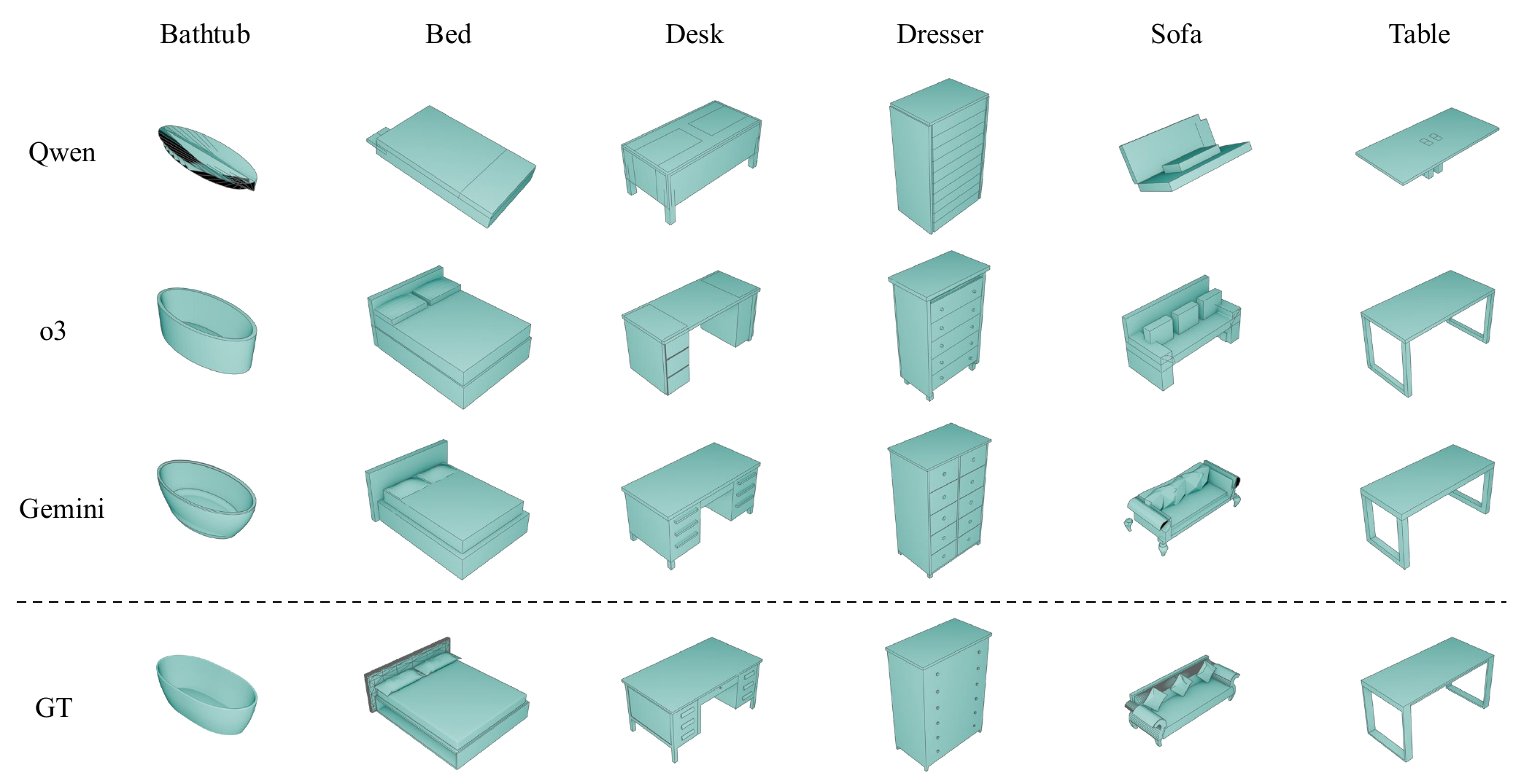}
    \caption{\textbf{Code-based reconstruction on ModelNet10.} 
All inputs come from the ModelNet10 dataset and contain only geometry; the cyan shading is for visualization only.}
    \label{fig:reconstruction_results}
    \vspace{-7pt}
\end{figure*}

\textbf{Code-based 3D Editing Paradigm.}
We edit programmatic 3D assets by taking a source script $\mathcal{S}_{src}$ and a text instruction $\mathcal{T}_{edit}$. The VLM parses $\mathcal{S}_{src}$, localizes the code relevant to $\mathcal{T}_{edit}$, and outputs an edited script $\mathcal{S}_{dst}$ with targeted changes.

\textbf{Construction of Localized 3D Editing Assets.}
\label{Dataset for Editing}
We build on the \textit{BlendNet} dataset~\citep{du2024blenderllmtraininglargelanguage} by selecting 55 representative objects and manually writing a high-quality and specific edit instruction for each. The resulting dataset, \textit{BlendNet-E}, contains triplets $(\text{source script}_i, \text{source description}_i, \text{edit instruction}_i)$.
The quantitative evaluation protocol for this task is presented in Section~\ref{sec:editing_eval}.

\section{Experiments}

\begin{table*}[t!]
\centering
\small
\setlength{\tabcolsep}{3pt}
\caption{\textbf{Reconstruction on ModelNet10.}
We report 3D metrics (CD, F@5\%, SBR@2\%, NC@5\%) and 2D metrics (Sil IoU, NRMSE, MAE); arrows indicate preferred directions. Metrics average successful outputs; open-source failure rates are reported in Appendix Table~\ref{tab:failrate-open}. Best and second-best values are highlighted in green and blue, within each comparison block.}
\label{tab:modelnet10}
\resizebox{.99\linewidth}{!}{
\renewcommand{\arraystretch}{1.1}
\begin{tabular}{lc *{4}{c} *{3}{c}}
\toprule
\multirow{3}{*}{Model} & \multirow{3}{*}{Paradigm} &
\multicolumn{7}{c}{\textbf{ModelNet10}}\\
& & \multicolumn{4}{c}{\textbf{3D Metrics}} & \multicolumn{3}{c}{\textbf{2D Metrics}} \\
\cmidrule(lr){3-6}\cmidrule(lr){7-9}
& & CD\,$\downarrow$ & F@5\%\,$\uparrow$ & SBR@2\%\,$\uparrow$ & NC@5\%\,$\uparrow$
& Sil IoU\,$\uparrow$ & NRMSE\,$\downarrow$ & MAE\,$\downarrow$
\\
\midrule
\multicolumn{9}{l}{\emph{Traditional mesh baselines}}\\
\addlinespace[2pt]
Unique3D~\citep{wu2024unique3d} & \NA & 0.0580 & 0.5679 & 0.2304 & 0.6236 & 0.5231 & 0.1388 & 0.4999 \\
\addlinespace[2pt]
InstantMesh~\citep{xu2024instantmesh} & \NA & \best{0.0191} & \best{0.9060} & \best{0.6443} & \best{0.8095} & \best{0.8068} & \best{0.0537} & \best{0.2110} \\
\addlinespace[2pt]\cdashline{1-9}[.5pt/2pt]\addlinespace[2pt]
\multicolumn{9}{l}{\emph{Code-based baseline}}\\
MeshCoder~\citep{dai2025meshcoder} & \NA & \second{0.0442} & \second{0.7053} & \second{0.4529} & \second{0.6544} & \second{0.6091} & \second{0.1250} & \second{0.3893} \\

\midrule
\addlinespace[3pt]
\multicolumn{9}{l}{\emph{Open-source VLM families}} \\
\addlinespace[2pt]
\multirow{4}{*}{LLaVA-OneVision-Qwen2-72B~\citep{li2024llavaonevisioneasyvisualtask}}
  & Sin. & 0.0674 & 0.5186 & 0.2603 & 0.5697 & \second{0.6018} & 0.1963 & 0.5460 \\
  & Pla. & 0.0704 & 0.5032 & 0.2757 & 0.5378 & 0.4753 & 0.1715 & 0.5188 \\
  & RAG & 0.0698 & 0.5137 & 0.2750 & 0.5838 & 0.4243 & 0.1504 & 0.4919 \\
  & Few. & 0.0699 & 0.4909 & 0.2756 & 0.5442 & 0.5093 & 0.1733 & 0.5423 \\
\addlinespace[2pt]\cdashline{1-9}[.5pt/2pt]\addlinespace[2pt]
\multirow{4}{*}{InternVL3.5-38B~\citep{wang2025internvl3_5}}
  & Sin. & 0.0673 & 0.5267 & 0.2614 & 0.5800 & 0.5236 & 0.1618 & 0.4984 \\
  & Pla. & 0.0642 & 0.5584 & 0.2956 & 0.5683 & 0.4983 & 0.1606 & 0.5048 \\
  & RAG & 0.0663 & 0.5458 & 0.2770 & 0.5721 & 0.4855 & 0.1492 & 0.4832 \\
  & Few. & 0.0648 & 0.5668 & 0.3152 & 0.5760 & 0.5326 & 0.1656 & 0.5158 \\
\addlinespace[2pt]\cdashline{1-9}[.5pt/2pt]\addlinespace[2pt]
\multirow{4}{*}{Qwen2.5-VL-72B-Instruct~\citep{qwen2.5-VL}}
  & Sin. & \second{0.0593} & \second{0.6038} & \second{0.3482} & 0.5849 & 0.5920 & 0.1558 & 0.4817 \\
  & Pla. & 0.0635 & 0.5995 & 0.3383 & \second{0.6024} & 0.5457 & 0.1458 & \second{0.4310} \\
  & RAG & 0.0665 & 0.5571 & 0.3004 & 0.5952 & 0.5067 & \second{0.1441} & 0.4421 \\
  & Few. & \best{0.0492} & \best{0.6702} & \best{0.4104} & \best{0.6197} & \best{0.6167} & \best{0.1289} & \best{0.4269} \\
\midrule
\addlinespace[3pt]
\multicolumn{9}{l}{\emph{Closed-source VLM families}} \\
\addlinespace[2pt]
\multirow{5}{*}{Claude Sonnet 4~\citep{anthropic_claude_sonnet4_2025}}
  & Sin. & 0.0405 & 0.7178 & 0.4435 & 0.6864 & 0.6333 & 0.1057 & 0.3307 \\
  & Pla. & 0.0419 & 0.7069 & 0.4541 & 0.6526 & 0.6264 & 0.1208 & 0.3733 \\
  & RAG & 0.0368 & 0.7394 & 0.4814 & 0.6641 & 0.6660 & 0.1171 & 0.3773 \\
  & Few. & 0.0374 & 0.7497 & 0.4818 & 0.6745 & 0.6561 & 0.1060 & 0.3491 \\
  & Agt. & 0.0435 & 0.7140 & 0.4990 & 0.6807 & 0.6262 & 0.1188 & 0.3652 \\
\addlinespace[2pt]\cdashline{1-9}[.5pt/2pt]\addlinespace[2pt]
\multirow{5}{*}{o3~\citep{openai_o3_2025}}
  & Sin. & 0.0608 & 0.5887 & 0.3185 & 0.6091 & 0.5469 & 0.1410 & 0.4340 \\
  & Pla. & 0.0328 & 0.7879 & 0.5312 & 0.6964 & 0.6921 & 0.0983 & 0.3085 \\
  & RAG & 0.0466 & 0.6642 & 0.3732 & 0.6536 & 0.5656 & 0.1221 & 0.3891 \\
  & Few. & 0.0318 & 0.7949 & 0.5306 & 0.7174 & 0.6934 & 0.0969 & 0.2939 \\
  & Agt. & 0.0510 & 0.6556 & 0.4058 & 0.6614 & 0.5540 & 0.1390 & 0.4006 \\
\addlinespace[2pt]\cdashline{1-9}[.5pt/2pt]\addlinespace[2pt]
\multirow{5}{*}{Gemini 3 Pro~\citep{google_gemini_3_pro_2025}}
  & Sin. & 0.0356 & 0.7624 & 0.4986 & 0.7134 & 0.6591 & 0.0998 & 0.3032 \\
  & Pla. & 0.0280 & 0.8277 & 0.5820 & 0.7190 & 0.7275 & 0.0900 & 0.2861 \\
  & RAG & 0.0251 & \best{0.8573} & \second{0.6199} & \second{0.7374} & \best{0.7585} & 0.0829 & \second{0.2681} \\
  & Few. & \second{0.0249} & \second{0.8532} & 0.6106 & \best{0.7427} & 0.7517 & \best{0.0748} & \best{0.2550} \\
  & Agt. & \best{0.0248} & 0.8513 & \best{0.6263} & 0.7332 & \second{0.7535} & \second{0.0826} & 0.2693 \\
\addlinespace[2pt]
\bottomrule
\end{tabular}
}
\vspace{-5pt}
\end{table*}

\subsection{3D Reconstruction Evaluation}
Our goal is to synthesize executable \texttt{bpy} code from a single input image and reconstruct a 3D object that matches the target as closely as possible. We evaluate this setting under the reconstruction benchmark described in Section~\ref{sec:recon_benchmark} and report results on ModelNet10 for four direct prompting paradigms, namely Single-call, Planning, RAG, and Few-shot, across both open-source and closed-source VLMs. We additionally study an Agent paradigm on the three closed-source models.

\subsubsection{Experimental Setup}

\paragraph{Datasets and Benchmark Setting.}
\label{Dataset for Reconstruction}
We use ModelNet10~\citep{modelnet_wu20153dshapenetsdeeprepresentation} under a controlled rendering protocol. Our reconstruction benchmark contains 100 objects sampled from ModelNet10 across its 10 categories. Each object is normalized and rendered from eight evenly spaced viewpoints on a sphere of radius 1.76, with depth and normal maps also generated for analysis. A human annotator selects the most informative RGB view as the single-image input. This selected view is fixed as the input condition image and as the camera view for 2D evaluation. To analyze sensitivity to structural complexity, we additionally split ModelNet10 into \emph{easy} and \emph{hard} subsets with one annotator and one verifier; details are provided in Section~\ref{ModelNet split}.

\paragraph{Models and Baselines.}

We evaluate code-based reconstruction on three open-source families, InternVL3.5 38B~\citep{wang2025internvl3_5}, LLaVA OneVision Qwen2 72B~\citep{li2024llavaonevisioneasyvisualtask}, and Qwen2.5 VL 72B Instruct~\citep{qwen2.5-VL}, and three closed-source families, Claude Sonnet 4~\citep{anthropic_claude_sonnet4_2025}, o3~\citep{openai_o3_2025}, and Gemini 3 Pro~\citep{google_gemini_3_pro_2025}. For all six models, we report Single-call, Planning, RAG, and Few-shot. For the three closed-source models, we additionally report Agent. We restrict Agent evaluation to the three closed-source models, where the multi-stage generation and self-correction workflow can be executed reliably in our current setup. We compare against Unique3D~\citep{wu2024unique3d}, InstantMesh~\citep{xu2024instantmesh}, and MeshCoder~\citep{dai2025meshcoder}. All VLM-based methods take a single RGB view as input, whereas MeshCoder uses ground-truth 3D point clouds.

\paragraph{Registration and Metrics.}
We follow the reconstruction benchmark in Section~\ref{sec:recon_benchmark} and apply the registration protocol in Section~\ref{sec:registration} before metric computation. We report four 3D metrics, Chamfer Distance (CD), F-score at 5\% (F@5\%), Spatial Balanced Recall at 2\% (SBR@2\%), and Normal Consistency at 5\% (NC@5\%), together with three input-view 2D metrics: Silhouette IoU (Sil IoU), depth NRMSE, and normal MAE. Detailed metric definitions are provided in Appendix~\ref{app:reconstruction_metrics}. For the external baselines and the closed-source VLM configurations emphasized in the main comparisons, we further quantify sampling uncertainty over object identities with 1,000-sample nonparametric bootstrap 95\% confidence intervals. These intervals are computed from evaluated object-level metric records and reported in Appendix Table~\ref{tab:modelnet10_bootstrap_ci}.

\paragraph{Reconstruction Variant.}
\label{Dataset for Reconstruction Variant}
To demonstrate the text-conditional reconstruction variant, we derive a companion test set from the same ModelNet10 assets and use the same human-selected informative views as inputs. For each image, we use GPT-4o~\citep{openai2024gpt4o} to generate a high-level edit instruction tailored to the depicted object, yielding 100 triplets of (image, edit instruction, source \texttt{.blend}). We denote this set by \textit{ModelNet10-V}. Since the set does not contain edited ground-truth shapes, we use it for qualitative analysis rather than quantitative reconstruction metrics.

\subsubsection{Main Results}
Table~\ref{tab:modelnet10} evaluates image-conditioned Blender-code reconstruction on ModelNet10 under four direct prompting paradigms and the Agent workflow, together with three external 3D reconstruction baselines. Representative reconstructions are shown in Figure~\ref{fig:reconstruction_results}, illustrating the geometry-only ModelNet10 setting used throughout this evaluation.

\textbf{Code-based reconstruction is competitive with several reference baselines under different input conditions.}
Gemini 3 Pro with Few-shot already outperforms Unique3D and MeshCoder on all reported metrics. For MeshCoder, it reduces CD from 0.0442 to 0.0249, and improves F@5\%, SBR@2\%, and NC@5\% from 0.7053/0.4529/0.6544 to 0.8532/0.6106/0.7427. This comparison is notable as all VLM-based methods use a single RGB view, while MeshCoder is conditioned on ground-truth point clouds. Compared with InstantMesh, Gemini 3 Pro with Agent narrows the gap on geometry-centric metrics, reaching CD 0.0248 and SBR@2\% 0.6263, compared with 0.0191 and 0.6443 for InstantMesh.

\textbf{Closed-source VLMs define the current frontier among code-based VLM reconstructions.}
Across the shared paradigms, closed-source VLMs define the frontier of code-based VLM performance. Under Few-shot, Gemini 3 Pro achieves CD 0.0249 and F@5\% 0.8532, while the best open-source counterpart, Qwen2.5-VL-72B-Instruct, reaches 0.0492 and 0.6702. The same separation is visible in SBR@2\% and 2D metrics. This pattern suggests that current reconstruction quality is more constrained by VLM capability than by the expressivity of executable code alone.

\textbf{Auxiliary structure interacts with model capability.}
The prompting paradigms do not form a universal ranking. Among open-source models, Qwen2.5-VL benefits most clearly from Few-shot prompting, reducing CD from 0.0593 to 0.0492 and improving F@5\% from 0.6038 to 0.6702 over Single-call, whereas LLaVA-OneVision and InternVL show less consistent gains. Closed-source models show different preferences: o3 benefits most from Planning and Few-shot, Claude Sonnet 4 is stronger with RAG/Few-shot than with Planning on most metrics, and Gemini 3 Pro remains strong under all structured paradigms. These comparisons suggest that plans, retrieved APIs, and exemplars provide complementary sources of structure rather than a fixed hierarchy.

\textbf{Agent favors part-balanced recovery.}
Agent adds explicit part-wise generation, intermediate execution, and final assembly. This emphasizes component coverage, which is directly reflected by SBR@2\%. Agent gives Claude Sonnet 4 its best SBR@2\%, and gives Gemini 3 Pro its best CD and SBR@2\%. It is not uniformly best because separately generated parts must still agree in scale, relative placement, and inter-part consistency. Such coordination errors can affect global surface agreement or visible-view alignment, explaining why Gemini 3 Pro with RAG obtains the best F@5\% and Sil IoU, Gemini 3 Pro with Few-shot obtains the best NC@5\%, NRMSE, and MAE, and o3-Agent remains below o3-Planning and o3-Few-shot on most metrics. Agent is therefore most useful when better part coverage outweighs the cost of multi-stage assembly.

\begin{figure}[t!]
    \centering
    
    \includegraphics[height=4.6cm]{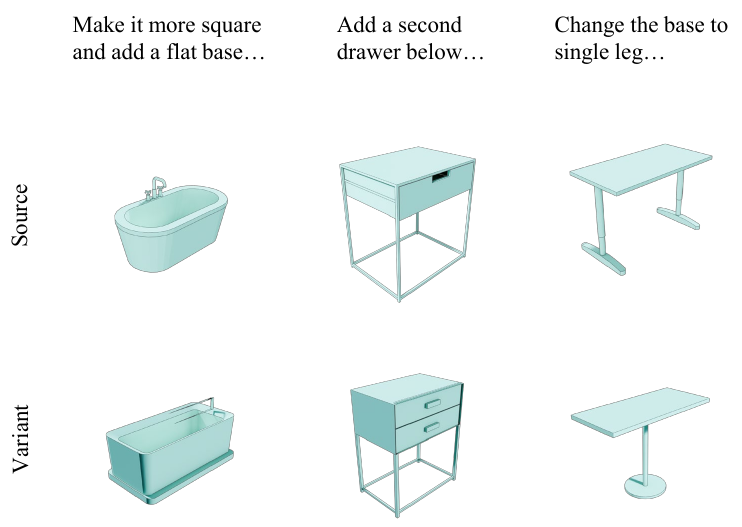}
    \hspace{0.2cm}
    \includegraphics[height=4.6cm]{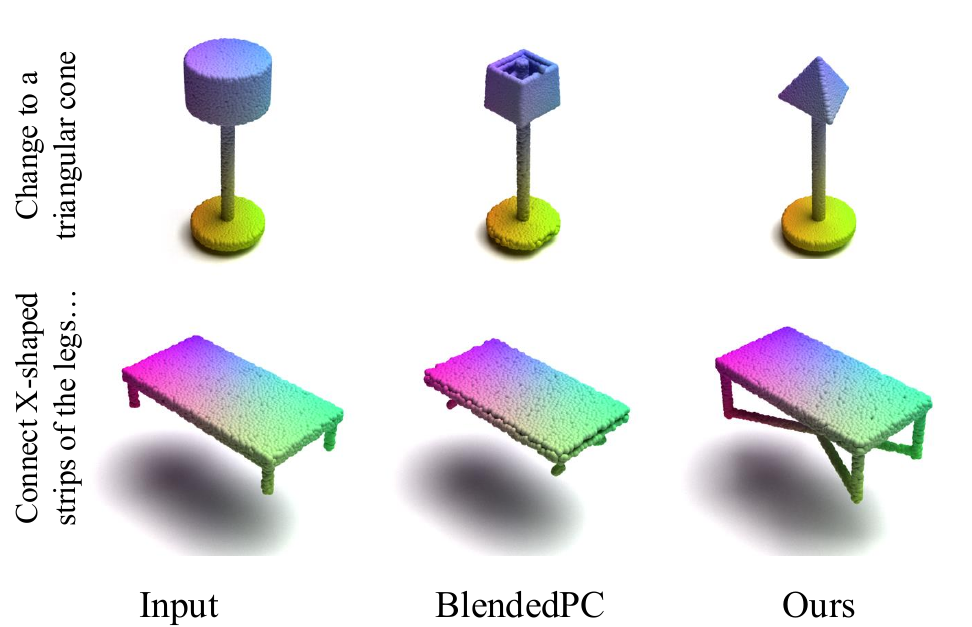}
\caption{\textbf{Reconstruction variant \& Code editing results.} 
(Left) Examples of our text-conditional reconstruction variant, which modifies an object based on a source image and a textual instruction. For the reconstruction variant, Source denotes the original object rather than an edited ground truth.
(Right) A direct comparison of our code editing method against the BlendedPC~\citep{sella2025blendedpc} baseline. Code-driven editing shows stronger edit fidelity while better preserving overall visual quality. 
A brief instruction is provided for each example in the figure; the full instructions are listed in Section~\ref{sec:editing_details}.}
    \label{fig:edit_point}
    \vspace{-7pt}
\end{figure}

\textbf{The easy/hard split supports the role of structural complexity.}
The easy/hard split captures structural complexity: easy objects have fewer parts, more regular structures, and milder curvature transitions, while hard objects contain more parts, irregular topology, or high-curvature transitions. The complete split in Appendix Table~\ref{tab:modelnet10_splits} includes the external 3D baselines and all evaluated closed-source VLM paradigms. Within that table, all five Gemini 3 Pro paradigms show higher Sil IoU and lower NRMSE/MAE on easy shapes, suggesting that current code-synthesis systems are most reliable when visual structure maps cleanly to parts, parameters, and transformations.

\textbf{Reconstruction variants as a bridge to editing.}
To demonstrate paradigm flexibility, we perform text-conditional reconstruction (Figure~\ref{fig:edit_point}, left), where models modify the source object based on textual instructions during generation. Using the o3 model on \textit{ModelNet10-V}, we qualitatively observe examples in which our approach interprets edit intents and applies targeted geometric modifications (additional examples in Figure~\ref{fig:variant_appendix}), suggesting the potential of code-based representations for semantic editing and motivating the following editing evaluation.

Overall, these results position executable code as a competitive representation for single-image 3D reconstruction, with the current frontier shaped by VLM capability and by structural workflows that turn visual evidence into reliable, editable programs.

\subsection{3D Code Editing Evaluation}
\label{sec:editing_eval}
We evaluate code editing on \textit{BlendNet-E}: given a source \texttt{bpy} script and a text instruction, the model edits the script to modify the asset. We compare with BlendedPC~\citep{sella2025blendedpc} on its suggested lamp/table subset and evaluate our o3-based method on the full dataset. Following BlendedPC's Edit Fidelity protocol, we render four orthogonal views and report average $\text{CLIP}_{sim}$ for text alignment and $\text{CLIP}_{dir}$, computed as in BlendedPC, for edit correctness.

\begin{table}[t]
\centering
\begin{minipage}[t]{0.55\linewidth}
\centering
\scalebox{0.85}{
\begin{tabular}{ccc}
  \toprule
  Method & $\text{CLIP}_{sim}\,\uparrow$ & $\text{CLIP}_{dir}\,\uparrow$ \\
  \midrule
  BlendedPC       & 0.0142 & 0.2017 \\
  $\text{Ours}_{P}$ & 0.0578 & 0.2499 \\
  $\text{Ours}_{A}$ & 0.0408 & 0.2469 \\
  \bottomrule
\end{tabular}
}
\captionof{table}{CLIP-based similarity and direction scores. $\text{Ours}_{P}$ is tested on the lamp and table categories compared with BlendedPC, while $\text{Ours}_{A}$ is tested on the entire \textit{BlendNet-E}.}
\label{tab:clip_metrics}
\end{minipage}
\hfill
\begin{minipage}[t]{0.38\linewidth}
\centering
\scalebox{0.9}{
\begin{tabular}{ccc}
  \toprule
  Method & Inst. $\,\uparrow$ & Pres. $\uparrow$ \\
  \midrule
  BlendedPC & 1.90 & 2.45 \\
  Ours & 4.37 & {4.30} \\
  \bottomrule
\end{tabular}
}
\captionof{table}{User study from 12 volunteers, scores (1--5) for BlendedPC vs. ours. Inst. = instruction following; Pres. = unedited preservation.}
\label{tab:user_study}
\end{minipage}
\vspace{-20pt}
\end{table}

\subsubsection{Main Results}
\textbf{Our method obtains stronger editing scores on the matched subset.} 
On the lamp/table subset, our code-based editing achieves $\text{CLIP}_{sim}=0.0578$ and $\text{CLIP}_{dir}=0.2499$, while BlendedPC records $0.0142$ and $0.2017$, respectively, as shown in Table~\ref{tab:clip_metrics} and in Figure~\ref{fig:edit_point}. This corresponds to a 4.1$\times$ higher score in $\text{CLIP}_{sim}$ and a +23.9\% gain in $\text{CLIP}_{dir}$, indicating both stronger text-image alignment and more faithful execution of the intended edit. 
In addition, a user study further favors our method in ``Instruction following" and ``Preservation of unedited regions", as shown in Table~\ref{tab:user_study}.

\noindent\textbf{Generalization to other categories.}
On the full \textit{BlendNet-E} set, $\text{Ours}_{A}$ retains a comparable $\text{CLIP}_{dir}$ but shows a lower $\text{CLIP}_{sim}$ than $\text{Ours}_{P}$, indicating consistent semantic editing across a broader set of shapes with weaker text-image similarity. We use $\text{Ours}_{P}$ for the matched lamp/table comparison against BlendedPC, while $\text{Ours}_{A}$ evaluates generalization beyond that subset.
Additional examples in Appendix Figures~\ref{fig:edit_result_all} and~\ref{fig:bnet_appendix} further show that code-based edits implement targeted geometric changes while preserving unmodified parts.

\subsection{Limitations}

Despite promising results, executable code is not yet a drop-in replacement for dedicated mesh reconstruction models: as shown in Table~\ref{tab:modelnet10}, our best Agent result narrows but does not close the gap to InstantMesh~\citep{xu2024instantmesh} on aggregate reconstruction metrics. Our experiments also show that part-wise Agent generation improves coverage but can introduce scale, placement, or inter-part inconsistencies; complex assemblies may still lead to missing dependencies, broken object relations, or non-executable code. Improving code-centric VLMs and their self-verification mechanisms is therefore important for making executable 3D representations both accurate and editable.

\section{Conclusion}

This work systematically evaluates executable Blender code as a representation for single-image 3D reconstruction. Across open- and closed-source VLMs, we study how prompting structure, retrieval, few-shot examples, and a component-level Agent workflow affect reconstruction quality. Results show that VLM-driven code synthesis recovers structured geometry, is competitive with the point-cloud-conditioned code baseline, and approaches representative mesh reconstruction baselines on selected metrics. We also show that code-based editing enables targeted semantic and geometric changes with stronger preservation than a point-cloud editing baseline, supporting executable code as a promising representation for editable 3D reconstruction.

\newpage
\bibliographystyle{plainnat}
\bibliography{ref}

\newpage
\appendix
\section{Appendix}

\subsection{Reconstruction Pipeline Details}

\paragraph{ModelNet10 Easy/Hard Split.}
\label{ModelNet split}
We partition {ModelNet10} into \emph{easy} and \emph{hard} subsets per category: objects with fewer parts, regular structures, and mild curvature transitions are labeled {easy}; objects with more parts, irregular topology, or pronounced/high-curvature transitions are labeled {hard}.
\begin{enumerate}[leftmargin=*,itemsep=2pt,topsep=2pt]
  \item \textbf{bathtub}: \\
    \emph{Easy}: bathtub\_0111, bathtub\_0119, bathtub\_0139, bathtub\_0154, bathtub\_0155; \\
    \emph{Hard}: bathtub\_0141, bathtub\_0153, bathtub\_0124, bathtub\_0115, bathtub\_0150.

  \item \textbf{Bed}: \\
    \emph{Easy}: bed\_0555, bed\_0557, bed\_0561, bed\_0572, bed\_0595; \\ 
    \emph{Hard}: bed\_0548, bed\_0566, bed\_0571, bed\_0614, bed\_0598.

  \item \textbf{Chair}: \\
    \emph{Easy}: chair\_0894, chair\_0897, chair\_0950, chair\_0901, chair\_0896; \\
    \emph{Hard}: chair\_0893, chair\_0891, chair\_0941, chair\_0898, chair\_0943.

  \item \textbf{Desk}: \\
    \emph{Easy}: desk\_0217, desk\_0262, desk\_0246, desk\_0236, desk\_0220; \\
    \emph{Hard}: desk\_0263, desk\_0253, desk\_0231, desk\_0209, desk\_0226.

  \item \textbf{Dresser}: \\
    \emph{Easy}: dresser\_0248, dresser\_0254, dresser\_0266, dresser\_0232, dresser\_0205; \\
    \emph{Hard}: dresser\_0209, dresser\_0257, dresser\_0243, dresser\_0217, dresser\_0233.

  \item \textbf{Monitor}: \\
    \emph{Easy}: monitor\_0503, monitor\_0545, monitor\_0535, monitor\_0531, monitor\_0528; \\
    \emph{Hard}: monitor\_0483, monitor\_0522, monitor\_0529, monitor\_0511, monitor\_0539.

  \item \textbf{night\_stand}: \\
    \emph{Easy}: night\_stand\_0207, night\_stand\_0232, night\_stand\_0263, night\_stand\_0231, night\_stand\_0283; \\
    \emph{Hard}: night\_stand\_0225, night\_stand\_0208, night\_stand\_0278, night\_stand\_0270, night\_stand\_0262.

  \item \textbf{Sofa}: \\
    \emph{Easy}: sofa\_0692, sofa\_0687, sofa\_0770, sofa\_0756, sofa\_0683; \\
    \emph{Hard}: sofa\_0761, sofa\_0777, sofa\_0745, sofa\_0746, sofa\_0743.

  \item \textbf{Table}: \\
    \emph{Easy}: table\_0443, table\_0439, table\_0399, table\_0422, table\_0447; \\
    \emph{Hard}: table\_0436, table\_0405, table\_0430, table\_0470, table\_0423.

  \item \textbf{Toilet}: \\
    \emph{Easy}: toilet\_0393, toilet\_0438, toilet\_0408, toilet\_0355, toilet\_0439; \\
    \emph{Hard}: toilet\_0419, toilet\_0409, toilet\_0367, toilet\_0436, toilet\_0401.
\end{enumerate}

\begin{table}[h]
\centering
\small
\setlength{\tabcolsep}{8pt}
\caption{\textbf{Open-source VLM failure rate} on ModelNet10. Values are \emph{fail/total}.}
\label{tab:failrate-open}
\begin{tabular}{lccc}
\toprule
\textbf{Strategy} & \textbf{InternVL3.5-38B} & \textbf{Qwen2.5-VL-72B} & \textbf{LLaVA-OneVision-72B} \\
\midrule
Single-call & 4/100 \fr{(4\%)}  & 4/100 \fr{(4\%)}  & 48/100 \fr{(48\%)} \\
Planning    & 27/100 \fr{(27\%)} & 3/100 \fr{(3\%)}  & 42/100 \fr{(42\%)} \\
RAG         & 24/100 \fr{(24\%)} & 7/100 \fr{(7\%)}  & 46/100 \fr{(46\%)} \\
Few-shot    & 28/100 \fr{(28\%)} & 1/100 \fr{(1\%)}  & 76/100 \fr{(76\%)} \\
\bottomrule
\end{tabular}
\end{table}
\paragraph{Failure rate of VLMs.}
We report the proportion of prompts that produced \emph{unsuccessful} runs on ModelNet10 for open-source VLMs. For each task, when the first generated code runs into an error, the model has 5 chances to correct it. If the code still reports an error after the chances are exhausted, the generation is considered to have failed. The failure rates are shown as fail/total in Table~\ref{tab:failrate-open}. For open-source models, we compute reconstruction metrics only on correctly generated samples. For closed-source models, failed initial generations are handled by the same automatic retry/self-correction protocol until an executable script is obtained; no manual code or mesh repair is used before evaluation.

\paragraph{Reconstruction Metrics.}
\label{app:reconstruction_metrics}
All 3D metrics are computed after the upright-preserving similarity alignment described in Section~\ref{sec:registration}. The evaluation samples surface points from the aligned generated mesh and the ground-truth mesh, and uses the ground-truth robust scale for normalization.

\textit{3D metrics.}
\begin{itemize}[leftmargin=*,itemsep=2pt,topsep=2pt]
    \item \textbf{Chamfer Distance (CD).} CD is the bidirectional nearest-neighbor surface distance, averaged over generated-to-ground-truth and ground-truth-to-generated directions and normalized by the ground-truth robust scale. Lower CD indicates better geometric accuracy.
    \item \textbf{F-score at 5\% (F@5\%).} F@5\% uses a threshold of 5\% of the ground-truth scale. Precision is the fraction of generated surface samples close to the ground truth, recall is the fraction of ground-truth samples recovered by the generated surface, and F@5\% is their harmonic mean. Higher values indicate better surface matching under this tolerance.
    \item \textbf{Spatial Balanced Recall at 2\% (SBR@2\%).} SBR@2\% measures ground-truth coverage under a 2\%-of-scale threshold after partitioning the ground-truth surface into spatial cells and averaging recall across occupied cells. Higher values indicate more spatially distributed recovery rather than recovery concentrated on large easy surfaces.
    \item \textbf{Normal Consistency at 5\% (NC@5\%).} NC@5\% averages the absolute dot product between matched surface normals for bidirectional nearest-neighbor pairs whose distance is within 5\% of the ground-truth scale. Higher values indicate better local surface orientation among matched regions.
\end{itemize}

For 2D evaluation, we use the same selected camera view as the single-image input and ray-cast the aligned generated mesh and the ground-truth mesh to obtain foreground masks, depths, and normals.

\textit{Input-view 2D metrics.}
\begin{itemize}[leftmargin=*,itemsep=2pt,topsep=2pt]
    \item \textbf{Silhouette IoU (Sil IoU).} Sil IoU is the intersection-over-union of the two foreground masks. Higher values indicate better projected shape agreement.
    \item \textbf{Depth NRMSE.} Depth NRMSE is the root mean squared depth error over the foreground intersection, normalized by the ground-truth robust scale. Lower values indicate better visible-depth recovery.
    \item \textbf{Normal MAE.} Normal MAE is the mean angular error between visible normals over the foreground intersection, normalized by $\pi/2$ and computed with an absolute normal dot product. Lower values indicate better visible surface-orientation agreement. If the foreground intersection is empty, depth NRMSE and normal MAE are set to 1.0 by the evaluation script.
\end{itemize}

\paragraph{Examples.}
The blueprint example is shown in Listing~\ref{supple:blueprint_example}, the Blender API example in Listing~\ref{supple:RAG database entry}, an example query generated by Gemini in Listing~\ref{supple:vlm_query}, and a retrieved RAG example in Listing~\ref{supple:rag_example}.

\begin{table*}[t]
\centering
\small
\setlength{\tabcolsep}{3pt}
\caption{\textbf{Quantitative reconstruction results on ModelNet10 \emph{easy}/\emph{hard} split.} ``Sin." stands for ``single-call", ``Pla." for ``planning", ``Few." for ``Few-shot", and ``Agt." for ``agent". Metrics: CD = Chamfer Distance, F@5\% = F-score at 5\% threshold, SBR@2\% = Spatial Balanced Recall at 2\% threshold, NC@5\% = Normal Consistency at 5\% threshold, Sil IoU = Silhouette IoU, NRMSE = Normalized RMSE, and MAE = Mean Angular Error.}
\label{tab:modelnet10_splits}
\resizebox{.99\linewidth}{!}{
\begin{tabular}{lc *{7}{c} *{7}{c}}
\toprule
\multirow{2}{*}{Model} & \multirow{2}{*}{Paradigm} &
\multicolumn{7}{c}{\textbf{ModelNet10-\emph{easy}}} &
\multicolumn{7}{c}{\textbf{ModelNet10-\emph{hard}}}\\
\cmidrule(lr){3-9}\cmidrule(lr){10-16}
& & CD\,$\downarrow$ & F@5\%\,$\uparrow$ & SBR@2\%\,$\uparrow$ & NC@5\%\,$\uparrow$ & Sil IoU\,$\uparrow$ & NRMSE\,$\downarrow$ & MAE\,$\downarrow$
& CD\,$\downarrow$ & F@5\%\,$\uparrow$ & SBR@2\%\,$\uparrow$ & NC@5\%\,$\uparrow$ & Sil IoU\,$\uparrow$ & NRMSE\,$\downarrow$ & MAE\,$\downarrow$ \\
\midrule
\multicolumn{16}{l}{\emph{External 3D baselines}}\\
\addlinespace[2pt]
Unique3D & \NA & 0.0589 & 0.5526 & 0.2236 & 0.6321 & 0.5225 & 0.1367 & 0.4952 & 0.0571 & 0.5832 & 0.2372 & 0.6151 & 0.5238 & 0.1408 & 0.5046 \\
\addlinespace[2pt]
InstantMesh & \NA & 0.0187 & 0.9066 & 0.6716 & 0.8253 & 0.8119 & 0.0484 & 0.1897 & 0.0194 & 0.9054 & 0.6169 & 0.7936 & 0.8017 & 0.0590 & 0.2324 \\
\addlinespace[2pt]
MeshCoder & \NA & 0.0464 & 0.6764 & 0.4438 & 0.6410 & 0.5868 & 0.1214 & 0.3806 & 0.0420 & 0.7354 & 0.4624 & 0.6685 & 0.6324 & 0.1288 & 0.3983 \\
\midrule
\addlinespace[3pt]
\multicolumn{16}{l}{\emph{Closed-source VLM families}} \\
\addlinespace[2pt]
\multirow{5}{*}{Claude Sonnet 4}
  & Sin. & 0.0394 & 0.7321 & 0.4952 & 0.7052 & 0.6653 & 0.1026 & 0.3055 & 0.0416 & 0.7036 & 0.3917 & 0.6675 & 0.6014 & 0.1088 & 0.3558 \\
  & Pla. & 0.0436 & 0.6940 & 0.4738 & 0.6500 & 0.6400 & 0.1216 & 0.3689 & 0.0403 & 0.7198 & 0.4345 & 0.6552 & 0.6129 & 0.1201 & 0.3777 \\
  & RAG & 0.0364 & 0.7380 & 0.4999 & 0.6706 & 0.6642 & 0.1088 & 0.3448 & 0.0372 & 0.7408 & 0.4629 & 0.6575 & 0.6678 & 0.1253 & 0.4099 \\
  & Few. & 0.0336 & 0.7734 & 0.5340 & 0.6977 & 0.6840 & 0.0929 & 0.3054 & 0.0413 & 0.7259 & 0.4296 & 0.6513 & 0.6283 & 0.1190 & 0.3929 \\
  & Agt. & 0.0422 & 0.7185 & 0.5059 & 0.6918 & 0.6412 & 0.1167 & 0.3451 & 0.0452 & 0.7085 & 0.4908 & 0.6673 & 0.6082 & 0.1214 & 0.3891 \\
\addlinespace[2pt]\cdashline{1-16}[.5pt/2pt]\addlinespace[2pt]
\multirow{5}{*}{o3}
  & Sin. & 0.0596 & 0.5906 & 0.3513 & 0.6253 & 0.5622 & 0.1363 & 0.4141 & 0.0620 & 0.5869 & 0.2857 & 0.5928 & 0.5317 & 0.1458 & 0.4538 \\
  & Pla. & 0.0317 & 0.7855 & 0.5455 & 0.7018 & 0.7049 & 0.0988 & 0.2950 & 0.0339 & 0.7902 & 0.5170 & 0.6911 & 0.6794 & 0.0978 & 0.3220 \\
  & RAG & 0.0432 & 0.6876 & 0.4157 & 0.6726 & 0.6055 & 0.1168 & 0.3580 & 0.0499 & 0.6402 & 0.3298 & 0.6341 & 0.5249 & 0.1274 & 0.4208 \\
  & Few. & 0.0308 & 0.7991 & 0.5517 & 0.7277 & 0.7082 & 0.0865 & 0.2654 & 0.0329 & 0.7907 & 0.5096 & 0.7071 & 0.6786 & 0.1072 & 0.3224 \\
  & Agt. & 0.0557 & 0.6245 & 0.3921 & 0.6371 & 0.5353 & 0.1394 & 0.4022 & 0.0460 & 0.6883 & 0.4201 & 0.6868 & 0.5736 & 0.1386 & 0.3990 \\
\addlinespace[2pt]\cdashline{1-16}[.5pt/2pt]\addlinespace[2pt]
\multirow{5}{*}{Gemini 3 Pro}
  & Sin. & 0.0346 & 0.7720 & 0.5350 & 0.7150 & 0.6926 & 0.0974 & 0.2991 & 0.0366 & 0.7528 & 0.4623 & 0.7117 & 0.6256 & 0.1023 & 0.3074 \\
  & Pla. & 0.0263 & 0.8444 & 0.6105 & 0.7273 & 0.7630 & 0.0820 & 0.2691 & 0.0297 & 0.8106 & 0.5530 & 0.7105 & 0.6912 & 0.0982 & 0.3035 \\
  & RAG & 0.0227 & 0.8710 & 0.6496 & 0.7478 & 0.7778 & 0.0735 & 0.2513 & 0.0275 & 0.8439 & 0.5908 & 0.7271 & 0.7395 & 0.0920 & 0.2845 \\
  & Few. & 0.0256 & 0.8502 & 0.6335 & 0.7500 & 0.7657 & 0.0678 & 0.2329 & 0.0241 & 0.8560 & 0.5881 & 0.7355 & 0.7380 & 0.0817 & 0.2766 \\
  & Agt. & 0.0239 & 0.8534 & 0.6318 & 0.7380 & 0.7701 & 0.0779 & 0.2519 & 0.0257 & 0.8492 & 0.6208 & 0.7285 & 0.7369 & 0.0874 & 0.2868 \\
\addlinespace[2pt]
\bottomrule
\end{tabular}
}
\end{table*}

\paragraph{Bootstrap uncertainty for selected reconstruction comparisons.}
We estimate sampling uncertainty over object identities using nonparametric bootstrap resampling. To keep the analysis focused, we report confidence intervals for the reconstruction comparisons most directly tied to the main conclusions: the external baselines used for reference comparisons and the closed-source VLM configurations emphasized in the main discussion. For each selected method and metric, we resample evaluated object-level metric records with replacement 1,000 times and recompute the mean. The 95\% confidence intervals are the 2.5th and 97.5th percentiles of the bootstrap distribution.
\begin{table*}[t]
\centering
\small
\setlength{\tabcolsep}{3pt}
\caption{\textbf{Object-level bootstrap 95\% confidence intervals for selected ModelNet10 reconstruction comparisons.}
We report mean [2.5th, 97.5th percentile] over 1,000 bootstrap resamples of evaluated object-level metric records.}
\label{tab:modelnet10_bootstrap_ci}
\resizebox{.99\linewidth}{!}{
\begin{tabular}{lcccc}
\toprule
\multicolumn{5}{c}{\textbf{3D Metrics}}\\
\midrule
\textbf{Method} & CD\,$\downarrow$ & F@5\%\,$\uparrow$ & SBR@2\%\,$\uparrow$ & NC@5\%\,$\uparrow$ \\
\midrule
InstantMesh & 0.0191 [0.0176, 0.0205] & 0.9060 [0.8898, 0.9213] & 0.6443 [0.6127, 0.6746] & 0.8095 [0.7950, 0.8242] \\
MeshCoder & 0.0442 [0.0380, 0.0507] & 0.7053 [0.6598, 0.7460] & 0.4529 [0.4049, 0.4999] & 0.6544 [0.6238, 0.6836] \\
Claude RAG & 0.0368 [0.0338, 0.0400] & 0.7394 [0.7105, 0.7668] & 0.4814 [0.4483, 0.5178] & 0.6641 [0.6429, 0.6855] \\
Claude Few-shot & 0.0374 [0.0340, 0.0410] & 0.7497 [0.7193, 0.7781] & 0.4818 [0.4418, 0.5219] & 0.6745 [0.6435, 0.7030] \\
o3 Few-shot & 0.0318 [0.0290, 0.0347] & 0.7949 [0.7652, 0.8222] & 0.5306 [0.4905, 0.5720] & 0.7174 [0.6970, 0.7378] \\
Gemini RAG & 0.0251 [0.0227, 0.0278] & 0.8573 [0.8336, 0.8797] & 0.6199 [0.5855, 0.6600] & 0.7374 [0.7173, 0.7559] \\
Gemini Few-shot & 0.0249 [0.0223, 0.0280] & 0.8532 [0.8327, 0.8740] & 0.6106 [0.5746, 0.6475] & 0.7427 [0.7257, 0.7605] \\
Gemini Agent & 0.0248 [0.0228, 0.0269] & 0.8513 [0.8301, 0.8706] & 0.6263 [0.5915, 0.6608] & 0.7332 [0.7162, 0.7508] \\
\bottomrule
\end{tabular}
}
\vspace{4pt}
\resizebox{.72\linewidth}{!}{
\begin{tabular}{lccc}
\toprule
\multicolumn{4}{c}{\textbf{2D Metrics}}\\
\midrule
\textbf{Method} & Sil IoU\,$\uparrow$ & NRMSE\,$\downarrow$ & MAE\,$\downarrow$ \\
\midrule
InstantMesh & 0.8068 [0.7875, 0.8254] & 0.0537 [0.0479, 0.0598] & 0.2110 [0.1930, 0.2283] \\
MeshCoder & 0.6091 [0.5669, 0.6497] & 0.1250 [0.1109, 0.1401] & 0.3893 [0.3485, 0.4273] \\
Claude RAG & 0.6660 [0.6362, 0.6994] & 0.1171 [0.1067, 0.1270] & 0.3773 [0.3437, 0.4093] \\
Claude Few-shot & 0.6561 [0.6240, 0.6901] & 0.1060 [0.0952, 0.1172] & 0.3491 [0.3108, 0.3905] \\
o3 Few-shot & 0.6934 [0.6609, 0.7247] & 0.0969 [0.0872, 0.1075] & 0.2939 [0.2676, 0.3206] \\
Gemini RAG & 0.7585 [0.7310, 0.7823] & 0.0829 [0.0723, 0.0929] & 0.2681 [0.2406, 0.2938] \\
Gemini Few-shot & 0.7517 [0.7255, 0.7784] & 0.0748 [0.0671, 0.0825] & 0.2550 [0.2300, 0.2821] \\
Gemini Agent & 0.7535 [0.7261, 0.7784] & 0.0826 [0.0745, 0.0914] & 0.2693 [0.2470, 0.2936] \\
\bottomrule
\end{tabular}
}
\end{table*}

\subsection{More Experimental Results}
\paragraph{Complete Results on ModelNet10-\emph{easy} and \emph{hard}.}
We evaluate the three closed-source VLMs on ModelNet10-\emph{easy} and \emph{hard} under the \emph{Single-call}, \emph{Planning}, \emph{RAG}, \emph{Few-shot}, and \emph{Agent} paradigms; results are summarized in Table~\ref{tab:modelnet10_splits}. The table also includes the external 3D baselines used in the main reconstruction comparison: Unique3D, InstantMesh, and MeshCoder. Overall, the split results are consistent with the main paper: code-based reconstruction generally performs better on easy shapes than on hard shapes, especially for the stronger Gemini variants, whereas the external 3D baselines show weaker or less consistent easy/hard trends. This suggests that current code-based reconstruction remains sensitive to structural complexity: easy objects map more cleanly to parts, parameters, and transformations, while hard objects require finer visual parsing and more precise spatial assembly.

\begin{figure}[t]
    \centering
    \includegraphics[width=0.75\linewidth]{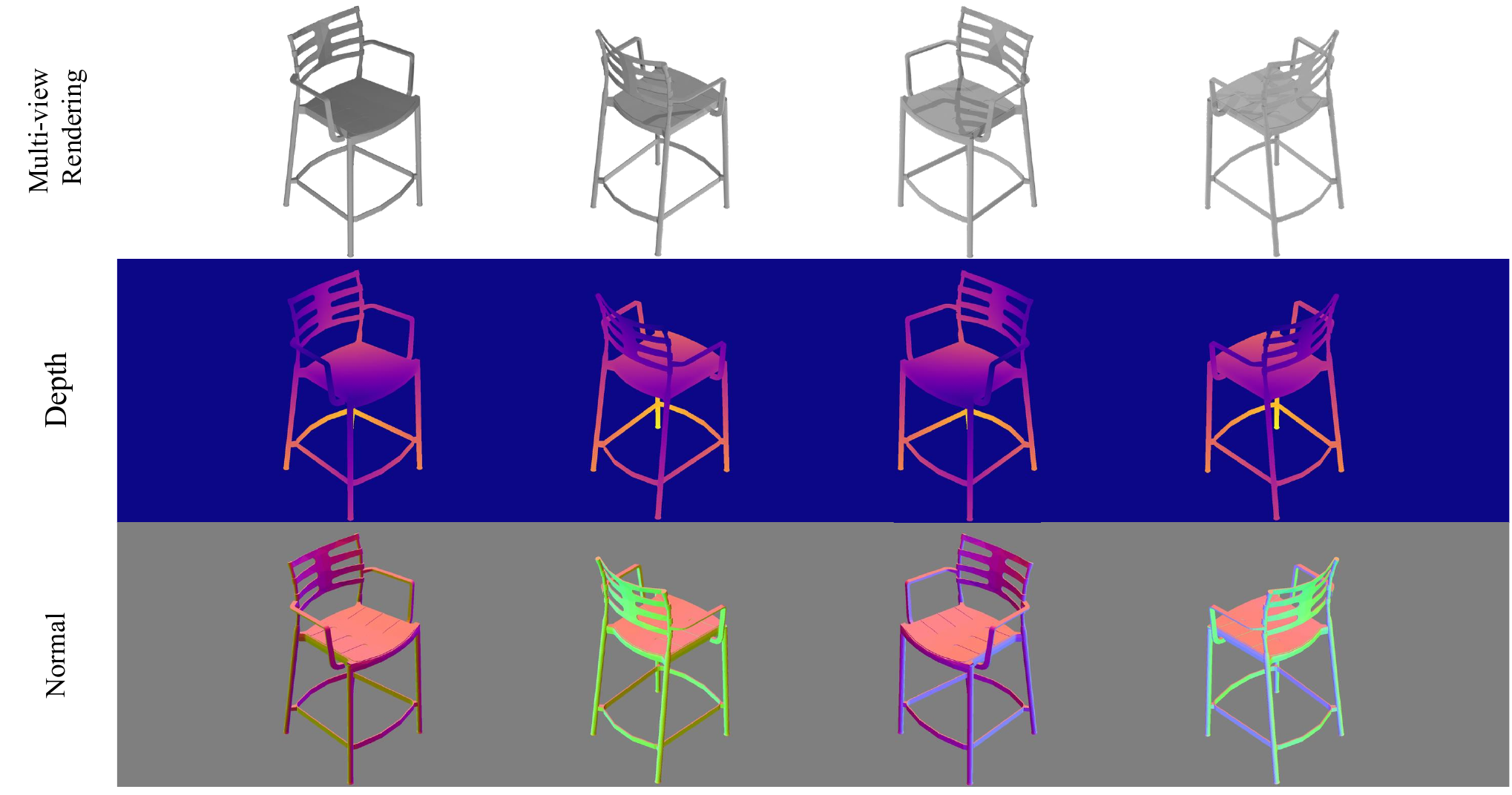}
    \caption{Rendered multi-view images of 3D object Chair 0891 in ModelNet10, with depth and surface normals.}
    \label{fig:depth_normal}
\end{figure}

\begin{figure}[t]
    \centering
    \includegraphics[width=0.95\linewidth]{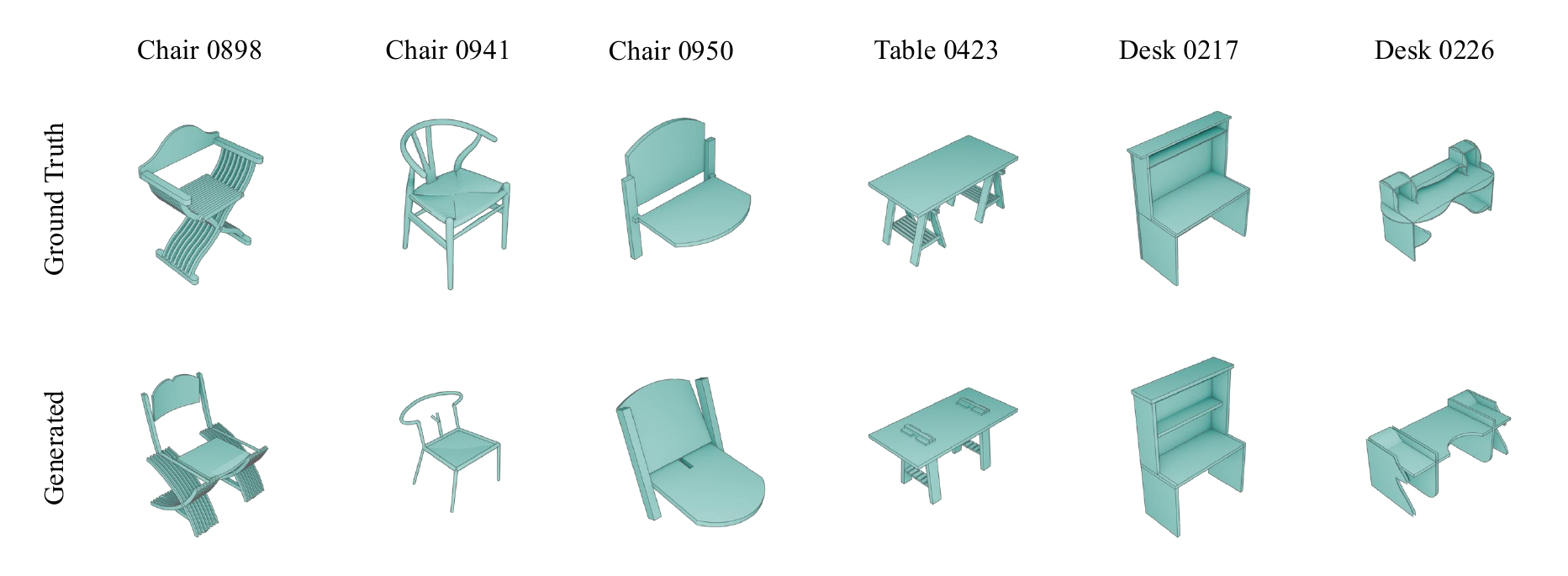}
    \caption{Failure cases demonstration. The generated objects preserve category-level semantics but deviate in fine geometric details.}
    \label{fig:gemini_bad_cases}
\end{figure}

\paragraph{Reconstruction Bad Cases Analysis.}
Figure~\ref{fig:gemini_bad_cases} demonstrates several failure cases of Gemini within the RAG paradigm. For Chair 0898, while the generated object exhibits a plausible shape and successfully produces chair legs with complex intersecting lines, it does not conform to the specifications of the ground truth. The reconstruction of Chair 0941 captures the general structure; however, the size of the "Y"-shaped backrest is incorrect, and the interconnecting components between the legs are missing. In Chair 0950, the individual components are generated approximately correctly, but their spatial arrangement is inaccurate, resulting in an overall structure that deviates significantly from the ground truth. At first glance, the object Table 0423 appears somewhat similar, but a detailed inspection reveals that the orientation of the legs and the angles of the connecting bars are rotated by 90 degrees. Furthermore, while the ground truth features four A-shaped leg structures, the generated object exhibits only two. For Desk 0217, the model misjudges the relative spacing, placing a horizontal bar at the midpoint instead of the correct position at one-quarter of the height. Desk 0226 possesses a complex structure with numerous curved elements and components. Although the final generated result bears a rough resemblance, the details differ substantially.

The primary failure modes can be summarized as follows: 
\begin{itemize}
    \item Incomplete comprehension of the input image, leading to missing components.
    \item Difficulty in accurately interpreting complex images, resulting in structures that are only coarsely similar to the ground truth.
    \item Insufficient spatial reasoning capability, causing failures in the correct assembly of components even when they are generated accurately.
\end{itemize}

\paragraph{Depth and Normal rendering examples.}
We render each object from multiple candidate viewpoints, as shown in Figure~\ref{fig:depth_normal}. A human-selected RGB view is used as the input condition image. For 2D metrics, we render both the prediction and the ground truth from this same camera view, so the evaluation quantifies recovery of the observed input view under the single-view condition.

\begin{table}[t]
\centering
\small
\setlength{\tabcolsep}{4pt}
\caption{Average token usage per item for closed-source model runs under different paradigms.}
\label{tab:tokens_cost}
\begin{tabular}{lccccc}
\toprule
& \textbf{single-call} & \textbf{planning} & \textbf{RAG} & \textbf{few-shot} & \textbf{Agent} \\
\midrule
Prompt tokens     & 1,222  & 9,093  & 11,867 & 9,800  & 118,000 \\
Completion tokens & 10,885 & 16,989 & 21,752 & 10,900 & 52,000  \\
Total tokens      & 12,107 & 26,082 & 33,619 & 20,700 & 170,000 \\
\bottomrule
\end{tabular}
\end{table}

\paragraph{Token cost.}
To better understand the practical overhead of different interaction paradigms, we report the average token usage per item for closed-source model runs in Table~\ref{tab:tokens_cost}. The single-call paradigm is the most economical. Planning, RAG, and few-shot prompting add structured context through blueprints, retrieved APIs, or exemplars, increasing the token footprint. The Agent paradigm incurs the largest token usage because it uses multi-stage planning, checking, part-wise code generation, and error correction; its prompt-token count includes cached-read tokens. These results highlight a clear trade-off between efficiency and performance. In practice, this suggests that single-call prompting may be preferable in resource-sensitive scenarios, whereas structured paradigms are more suitable when accuracy or part-level fidelity is prioritized and additional token overhead is acceptable.

\begin{figure}[t]
    \centering
    \includegraphics[width=0.95\linewidth]{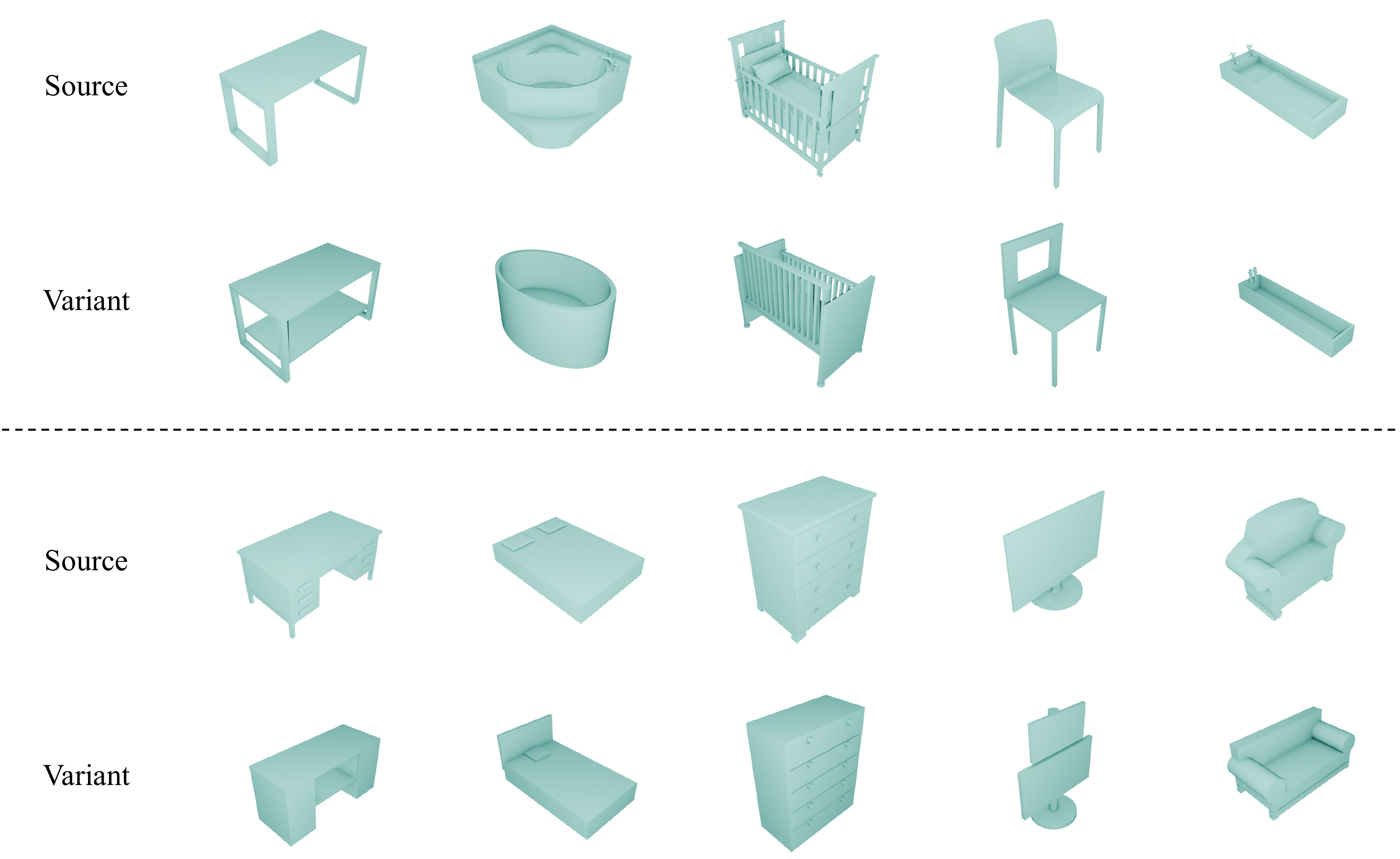}
    \caption{\textbf{Additional examples of the text-conditional reconstruction variant.} The model generates a modified 3D asset based on a source image from ModelNet10 and a corresponding text instruction. Here Source denotes the original object rather than an edited ground truth.}
    \label{fig:variant_appendix}
\end{figure}

\paragraph{Additional Qualitative Results.}
We provide additional qualitative examples for the \textbf{text-conditional reconstruction variant} in Figure~\ref{fig:variant_appendix} and for \textbf{code editing} in Figures~\ref{fig:edit_result_all} and~\ref{fig:bnet_appendix}. The complete editing instructions are listed in Section~\ref{sec:editing_details}.

\begin{figure}[t]
    \centering
    \includegraphics[width=0.5\linewidth]{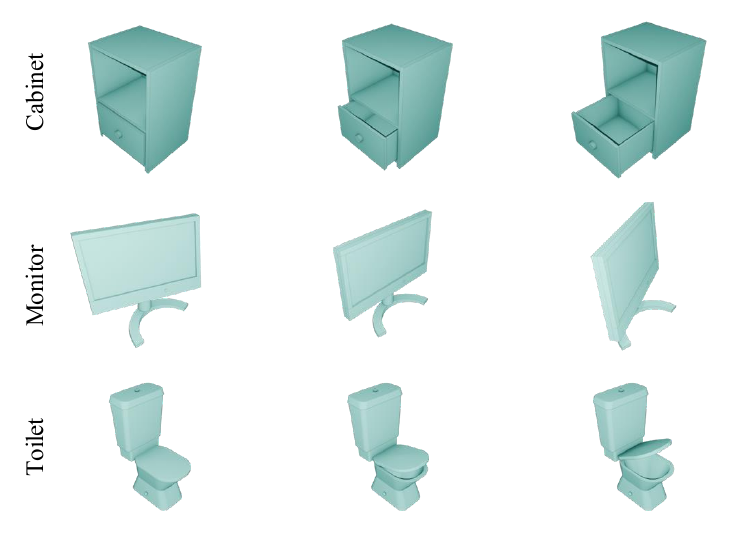}
    \caption{\textbf{Visualization of articulated objects.} We demonstrate the articulation effects generated by our Blender script across three categories. These include the sliding translation of cabinet drawers, the multi-axis rotation of monitor screens, and the hinge-based rotation of toilet lids, all simulating real-world usage scenarios.}
    \label{fig:articulated_samples}
\end{figure}

\paragraph{Articulation Results.}
We evaluate the effectiveness of our Blender-based articulation pipeline on samples from three object categories: Cabinet, Monitor, and Toilet. Specifically, we utilize shape keys to implement the translational motion of cabinet drawers, while applying rotational transformations to achieve the horizontal and vertical swiveling of monitor screens and the axial rotation of toilet lids. The generated motions are designed to follow plausible physical motion patterns. The qualitative results are illustrated in Figure~\ref{fig:articulated_samples}.

\subsection{Editing pipeline details}
\label{sec:editing_details}

\begin{figure}[t]
    \centering
    \includegraphics[width=0.95\linewidth]{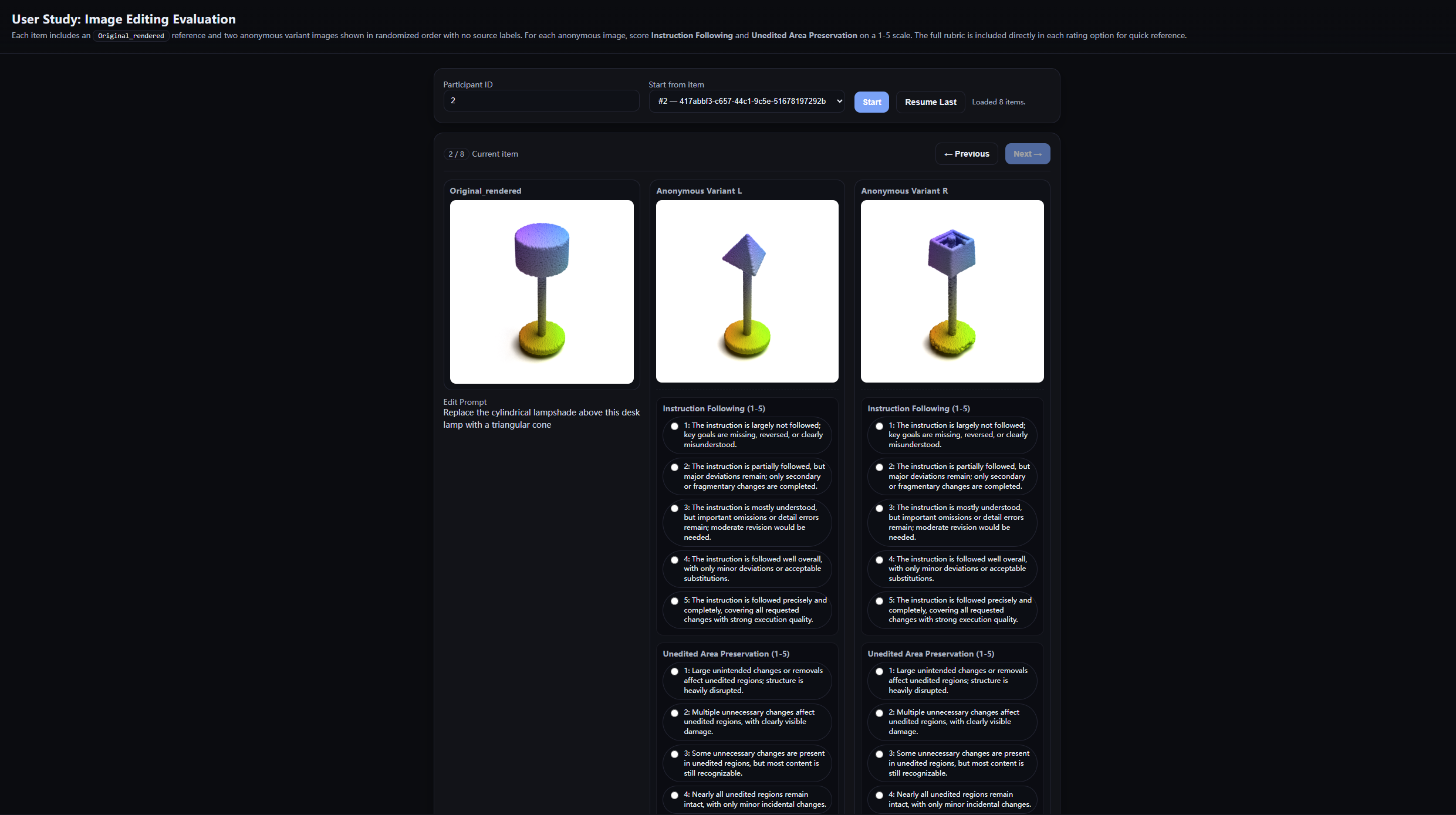}
    \caption{\textbf{User study interface for image editing evaluation.} Each item presents one reference render, two anonymous edited variants in randomized left/right order, the edit prompt, and rating controls for instruction following and preservation of unedited regions.}
    \label{fig:user_study_interface}
\end{figure}

\paragraph{User study interface.}
Figure~\ref{fig:user_study_interface} shows the interface used in our user study for comparing anonymous edited results against the same reference image and edit prompt.

\paragraph{Complete Instructions Used for Editing.}
For visual clarity, we only show the most critical text instruction when showing the edit figures, and omit the long part with \texttt{...}. Here we list the complete instructions.

\begin{figure}[t]
    \centering
    \includegraphics[width=0.95\linewidth]{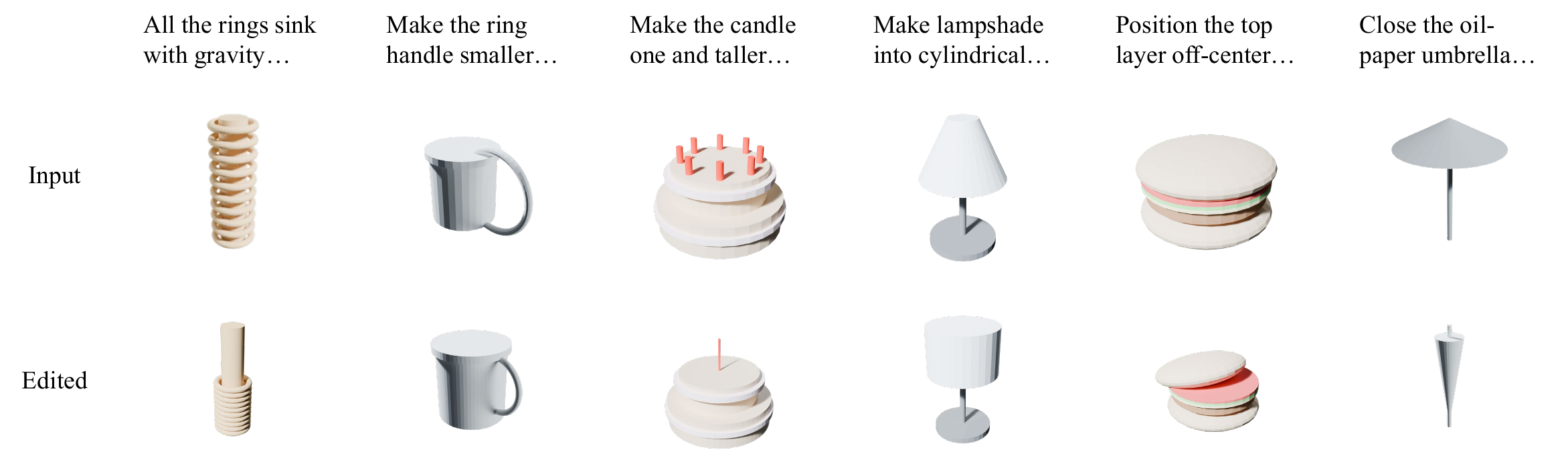}
    \caption{\textbf{Qualitative results for code editing.} Each example shows an input and edited object from \textit{BlendNet-E} with a shortened text instruction; complete instructions are listed below.}
    \label{fig:edit_result_all}
\end{figure}

In Figure~\ref{fig:edit_result_all}, the instructions we use are:

\begin{enumerate}
    \item Let all the rings on the pillar sink with gravity and fit together.
    \item The ring handle on the side of this cup is too big and does not match the cup body. Make it smaller.
    \item The candle on this cake is too thick and short. Make it thinner and longer and reduce the number to 1 and insert it in the middle of the top of the cake.
    \item Change the frustum-shaped lampshade of the upper part of the table lamp into a cylindrical shape.
    \item Position the top layer of the burger off-center so people can see the insides.
    \item This oil-paper umbrella is open, with a cone on top. Turn it to the closed position.
\end{enumerate}

In Figure~\ref{fig:edit_point}, the instructions we use are:
\begin{enumerate}
    \item Make the bathtub more square and add a flat base for stability.
    \item Add a second drawer below the existing one.
    \item Change the base legs to a single centered pedestal.
    \item Replace the cylindrical lampshade above this desk lamp with a triangular cone.
    \item Make this table lamp taller. The column mistakenly passes through the lampshade and protrudes a little from the top. Remove this small part.
    \item Lengthen the four cylindrical legs of this table and connect the legs at opposite corners at the bottom with X-shaped wooden strips to make its structure more stable.
\end{enumerate}

\begin{figure}[t]
    \centering
    \includegraphics[width=0.95\linewidth]{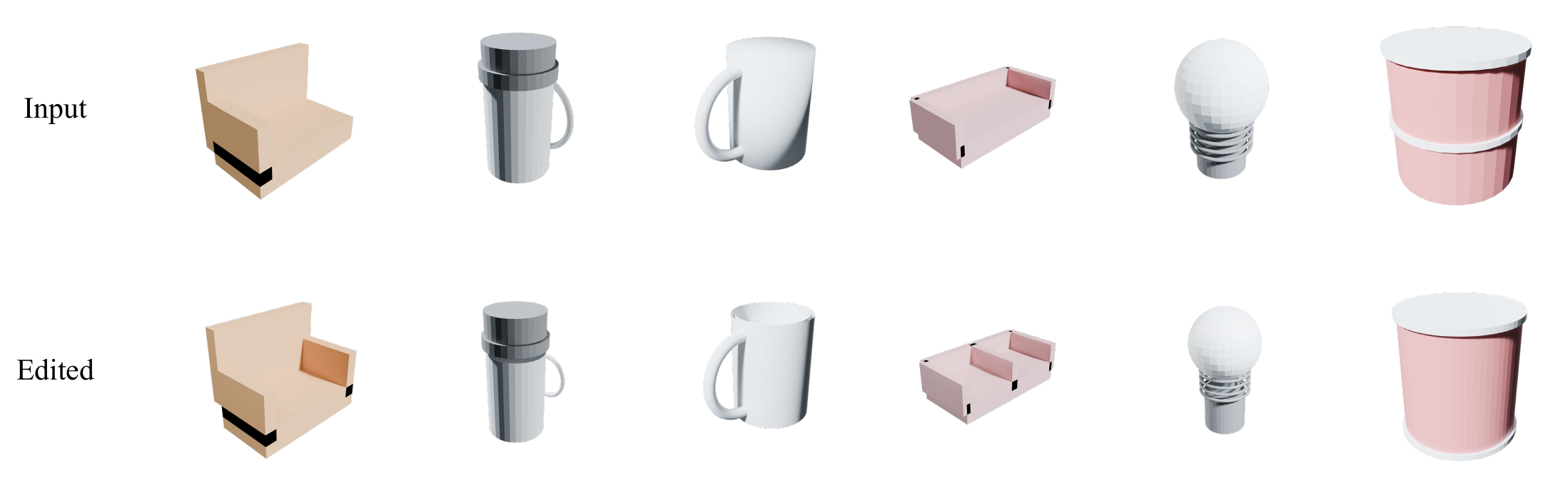}
    \caption{\textbf{Complementary code editing examples on BlendNet-E.} These examples cover additional objects and edit intents beyond Figure~\ref{fig:edit_result_all}.}
    \label{fig:bnet_appendix}
\end{figure}

In Figure~\ref{fig:variant_appendix}, the instructions we use are:
\begin{itemize}
    \item Upper part:
        \begin{enumerate}
            \item Add a lower shelf between the two legs.
            \item Convert the corner bath to an oval shape.
            \item Convert one of the crib's sides into a removable panel.
            \item Cut a large opening in the middle of the backrest.
            \item Extend the basin to double its current length.
        \end{enumerate}
    \item Lower part:
        \begin{enumerate}
            \item Add a central open shelf in the knee space area for additional storage.
            \item Add a headboard to the bed.
            \item Add a fifth drawer at the bottom.
            \item Add a second, smaller screen on top to create a dual-monitor setup.
            \item Add a lower central support beam between the sofa legs.
        \end{enumerate}
\end{itemize}

In Figure~\ref{fig:bnet_appendix}, the instructions we use are:
\begin{enumerate}
    \item This sofa has armrests on only one side and the modification makes it have armrests on both sides.
    \item The keychain circle on this cup is too big; make it smaller.
    \item The cylindrical portion of this cup was incorrectly generated as a solid shape. Make it hollow.
    \item Add a handguard in the middle of this sofa to give it two separate seats.
    \item Separate the spherical part of this bulb from the base.
    \item This bucket has an ugly ring around the cylindrical waist. Remove it.
\end{enumerate}

\newpage

\subsection{Detailed Pipeline Artifacts}
\label{sec:detail_artifacts}

\begin{figure}[h!]
  \centering
  \begin{minipage}{0.95\linewidth}
  \lstset{
    basicstyle=\ttfamily\scriptsize,
    breaklines=true,
    breakatwhitespace=true,
    columns=flexible,
    frame=tb,
    keepspaces=true,
    showstringspaces=false
  }
\begin{lstlisting}[caption={Blueprint JSON for object in Figure~\ref{fig:reconstruction_pipeline}. The blueprint normalizes to a base dimension (\texttt{overall\_height}=1.0) and encodes part-wise parametrics used by the code generator.},label={supple:blueprint_example}]
{
  "object_category": "bar_stool",
  "base_dimension": "overall_height",
  "dimensional_profile": {
    "overall_height": 1.0,
    "overall_width_at_base_ratio_to_height": 0.42,
    "overall_depth_at_base_ratio_to_height": 0.4,
    "seat_height_from_ground_ratio_to_overall_height": 0.65
  },
  "geometric_components": {
    "legs": {
      "count": 4,
      "profile_shape": "cylindrical",
      "diameter_ratio_to_overall_height": 0.02,
      "splay_angle_from_vertical_degrees": 6.0
    },
    "footrest": {
      "structure_type": "continuous_four_sided_brace",
      "height_from_ground_ratio_to_overall_height": 0.18,
      "cross_section_diameter_ratio_to_leg_diameter": 1.0,
      "front_bar_outward_curve_depth_ratio_to_overall_depth": 0.15
    },
    "seat": {
      "plan_shape": "rounded_square",
      "width_ratio_to_overall_width_at_base": 0.86,
      "depth_ratio_to_overall_depth_at_base": 0.85,
      "thickness_ratio_to_overall_height": 0.03,
      "ergonomic_concave_dip_ratio_to_seat_depth": 0.05,
      "front_edge_waterfall_radius_ratio_to_seat_thickness": 1.0,
      "cutouts": {
        "count": 2,
        "type": "slot",
        "slot_length_ratio_to_seat_depth": 0.7,
        "slot_width_ratio_to_seat_width": 0.08,
        "slot_spacing_from_centerline_ratio_to_seat_width": 0.25
      }
    },
    "backrest": {
      "height_above_seat_ratio_to_overall_height": 0.35,
      "width_ratio_to_seat_width": 0.95,
      "tilt_angle_from_vertical_degrees": 12.0,
      "horizontal_lumbar_curve_depth_ratio_to_width": 0.08,
      "structure": {
        "type": "slatted_frame",
        "frame_thickness_ratio_to_leg_diameter": 1.2,
        "slat_count": 3,
        "slat_height_ratio_to_backrest_height": 0.12,
        "slat_vertical_gap_ratio_to_slat_height": 1.1,
        "vertical_support_count": 2,
        "vertical_support_width_ratio_to_frame_thickness": 1.0
      }
    },
    "armrests": {
      "count": 2,
      "structure_type": "continuous_loop_from_backrest_to_seat",
      "height_above_seat_at_rear_ratio_to_overall_height": 0.15,
      "length_ratio_to_seat_depth": 0.8,
      "cross_section_diameter_ratio_to_leg_diameter": 1.0,
      "downward_slope_angle_degrees": 3.0,
      "outward_bow_distance_ratio_to_seat_width": 0.05
    }
  }
}
\end{lstlisting}
\end{minipage}
\end{figure}

\newpage
\begin{figure}[t]
  \centering
  \begin{minipage}{0.95\linewidth}
  \lstset{
    basicstyle=\ttfamily\scriptsize,
    breaklines=true,
    breakatwhitespace=true,
    columns=flexible,
    frame=tb,
    keepspaces=true,
    showstringspaces=false
  }
\begin{lstlisting}[caption={Blender 4.4 Python API database entry (structured JSON).},label={supple:RAG database entry}]
{
  "symbol": "bpy.ops.curves.add_bezier",
  "language": "python",
  "module": "bpy.ops.curves",
  "signature": ".ops.curves.add_bezier(*, radius=1.0, enter_editmode=False, align='WORLD', location=(0.0, 0.0, 0.0), rotation=(0.0, 0.0, 0.0), scale=(0.0, 0.0, 0.0))",
  "parameters": [
    {"name": "radius", "description": "radius (float in [0, inf], optional) - Radius"},
    {"name": "enter_editmode", "description": "enter_editmode (boolean, optional) - Enter Edit Mode when adding this object"},
    {"name": "align", "description": "align (enum in ['WORLD', 'VIEW', 'CURSOR'], optional) - Alignment of the new object. WORLD: align to world. VIEW: align to view. CURSOR: use 3D cursor orientation."},
    {"name": "location", "description": "location (mathutils.Vector of 3 items in [-inf, inf], optional) - Location for the newly added object"},
    {"name": "rotation", "description": "rotation (mathutils.Euler of 3 items in [-inf, inf], optional) - Rotation for the newly added object"},
    {"name": "scale", "description": "scale (mathutils.Vector of 3 items in [-inf, inf], optional) - Scale for the newly added object"}
  ],
  "doc_text": "Add new bezier curve",
  "version": "4.4"
}
\end{lstlisting}
\end{minipage}
\end{figure}

\newpage
\begin{figure}[t]
\centering
\begin{minipage}{0.95\linewidth}
  \lstset{
    basicstyle=\ttfamily\scriptsize,
    breaklines=true,
    breakatwhitespace=true,
    columns=flexible,
    frame=tb,
    keepspaces=true,
    showstringspaces=false
  }
\begin{lstlisting}[caption={VLM-generated intent queries per blueprint component.},label={supple:vlm_query}]
{
  "intent_queries": {
    "queries": {
      "legs": {
        "prefer_modules": ["bpy.ops.mesh", "bpy.ops.object"],
        "keywords": ["primitive_cylinder_add", "cylinder", "duplicate_move", "rotate", "splay"],
        "query": "Create four cylindrical legs for a bar stool. Start by adding a cylinder primitive, then duplicate it and rotate the legs to create a splayed angle from the vertical."
      },
      "footrest": {
        "prefer_modules": ["bpy.ops.curve", "bpy.ops.object"],
        "keywords": ["curve", "bezier", "bevel", "extrude", "depth", "join"],
        "query": "Model a continuous four-sided footrest brace with a curved front. Use a bezier curve with bevel depth to form the cylindrical rails and join the segments."
      },
      "seat": {
        "prefer_modules": ["bpy.ops.mesh", "bpy.ops.object"],
        "keywords": ["primitive_cube_add", "subdivision_set", "bevel", "boolean", "proportional_edit", "loop_cut"],
        "query": "Create a rounded square seat with an ergonomic dip and two slot cutouts. Start with a cube, use subdivision and proportional editing for the dip, bevel the edges, and apply a boolean difference modifier for the slots."
      },
      "backrest": {
        "prefer_modules": ["bpy.ops.mesh", "bpy.ops.object"],
        "keywords": ["plane", "extrude", "inset", "boolean", "loop_cut", "curve_deform", "modifier"],
        "query": "Create a slatted backrest frame that is tilted and curved. Model the basic shape from a plane using extrude and inset, use a boolean modifier to create the slats, and bend the result with a curve deform modifier."
      },
      "armrests": {
        "prefer_modules": ["bpy.ops.curve", "bpy.ops.object"],
        "keywords": ["curve", "bezier", "extrude", "bevel", "depth", "mirror"],
        "query": "Create two continuous loop armrests extending from the backrest to the seat. Model one armrest using a bezier curve with a circular bevel depth, then use a mirror modifier to create the second one."
      }
    }
  }
}
\end{lstlisting}
\end{minipage}
\end{figure}

\newpage
\begin{figure}[t]
  \centering
  \begin{minipage}{0.95\linewidth}
  \lstset{
    basicstyle=\ttfamily\scriptsize,
    breaklines=true,
    breakatwhitespace=true,
    columns=flexible,
    frame=tb,
    keepspaces=true,
    showstringspaces=false
  }
\begin{lstlisting}[caption={Retrieved APIs organized by component (post-retrieval structuring).},label={supple:rag_example}]
{
  "components": [
    {
      "name": "legs",
      "apis": [
        {
          "symbol": "bpy.ops.mesh.primitive_cylinder_add",
          "signature": "primitive_cylinder_add(vertices=32, radius=1.0, depth=2.0, end_fill_type='NGON', calc_uvs=True, enter_editmode=False, align='WORLD', location=(0.0, 0.0, 0.0), rotation=(0.0, 0.0, 0.0), scale=(0.0, 0.0, 0.0))",
          "parameters": {
            "vertices": "int in [3, 10000000], optional - Number of vertices for the circular caps.",
            "radius": "float in [0, inf], optional - Radius of the cylinder.",
            "depth": "float in [0, inf], optional - Height of the cylinder.",
            "end_fill_type": "enum in ['NOTHING', 'NGON', 'TRIFAN'], optional - How to fill the cylinder ends.",
            "location": "3D vector, optional - Location for the newly added object.",
            "rotation": "3D Euler rotation, optional - Rotation for the newly added object.",
            "scale": "3D vector, optional - Scale for the newly added object."
          }
        }
      ]
    },
    { "name": "footrest", "apis": [] },
    { "name": "seat", "apis": [] },
    { "name": "backrest", "apis": [] },
    {
      "name": "armrests",
      "apis": [
        {
          "symbol": "bpy.ops.curves.add_bezier",
          "signature": "add_bezier(radius=1.0, enter_editmode=False, align='WORLD', location=(0.0, 0.0, 0.0), rotation=(0.0, 0.0, 0.0), scale=(0.0, 0.0, 0.0))",
          "parameters": {
            "radius": "float in [0, inf], optional - Radius to set for the curve points.",
            "enter_editmode": "boolean, optional - Enter Edit Mode after creating the object.",
            "align": "enum in ['WORLD', 'VIEW', 'CURSOR'], optional - Alignment of the new object. WORLD: align to world. VIEW: align to view. CURSOR: use 3D cursor orientation."
          }
        }
      ]
    }
  ]
}
\end{lstlisting}
\end{minipage}
\end{figure}

\end{document}